\documentclass[10pt]{article}
\usepackage{todonotes}
\usepackage[accepted]{tmlr}
\usepackage{wrapfig}
\usepackage{graphicx}
\usepackage{booktabs}
\usepackage{xcolor}
\usepackage{subcaption}
\usepackage{bbm}
\usepackage{multirow}
\usepackage[table]{xcolor} 
\usepackage{float}
\usepackage{enumitem}
\usepackage{makecell}
\usepackage{microtype}
\usepackage[accsupp]{axessibility}  

\usepackage{amsmath}
\usepackage{amsthm}
\def\month{09}
\def\year{2026}
\def\openreview{\url{https://openreview.net/forum?id=E8oetPdxH6}}

\newtheorem{theorem}{Theorem}

\theoremstyle{definition}
\newtheorem{definition}[theorem]{Definition}

\theoremstyle{remark}

\usepackage{hyperref}

\usepackage{orcidlink}

\newcommand{\x}{\mathbf{x}}

\newcommand{\bc}{\mathbf{c}}

\definecolor{chudiorange}{RGB}{230,120,20}

\begin{document}

\title{Parameter-Efficient Construction of the Rashomon Slice for Concept Bottleneck Models} 
\author{\name Shihan Feng\thanks{Equal contribution.} \email{shihanf@unc.edu}\\
\addr University of North Carolina at Chapel Hill
\AND
\name Cheng Zhang\footnotemark[1] \email{zch@unc.edu} \\
\addr University of North Carolina at Chapel Hill
\AND
\name Michael Xi 
\email{michael.xi@rutgers.edu}\\
\addr Rutgers University
\AND
\name Ethan Hsu 
\email{ethan.hsu.contact@gmail.com} \\
\addr Duke University
\AND
\name Lesia Semenova 
\email{lesia.semenova@rutgers.edu}\\
\addr Rutgers University
\AND
\name Chudi Zhong 
\email{chudi@unc.edu}\\
\addr University of North Carolina at Chapel Hill
}

\maketitle

\begin{abstract}
In many machine learning problems, there may exist multiple models that achieve nearly identical predictive performance while relying on fundamentally different internal logic. 
However, standard training procedures produce a single model, offering no practical way to explore alternatives that may better suit downstream needs.
The set of these equally accurate models is known as the Rashomon set. Exploring the Rashomon set is particularly challenging in large and complex hypothesis spaces, such as Concept Bottleneck Models (CBMs), which are widely used in computer vision to make predictions through intermediate, human-understandable concepts.
In this paper, we provide a method for efficiently exploring the Rashomon set of CBMs. Our framework introduces a specialized parallel parameter-efficient adaptation module, combined with a checkpointing scheme and a concept diversity objective, to generate multiple equally accurate CBMs from a single training process. Empirical results show that our method finds models with better diversity than baselines while using much less memory. We further demonstrate that access to these diverse yet accurate CBMs enables trustworthy model selection, resolution of inter-class confusion, and reliable abstention in decision-making.

\end{abstract}

\section{Introduction}
\label{sec:intro}

Interpretability is important for deploying machine learning models in high-stakes domains, as it can help practitioners to troubleshoot errors, audit for biases, and align model behavior with domain expertise. For image classification, concept bottleneck models (CBMs) are designed to be interpretable by first predicting human-interpretable concepts and then using those concepts for the final classification. However, current CBM frameworks operate under a ``unique truth'' paradigm: they are optimized to produce a single model. This paradigm is fundamentally limited; if the resulting model relies on a flawed rationale or a spurious correlation, practitioners have no immediate alternative ``good'' models to turn to. 

Indeed, in many machine learning problems, there could be different models that achieve similar predictive accuracy yet differ substantially in reasoning patterns, known as \textit{the Rashomon set}  \citep{SemenovaRuPa2022, RudinEtAlAmazing2024, xin2022exploring}.
For example, a bird species may be identified with equal accuracy by its wing pattern or its beak shape. This diversity in concept-based reasoning rationale is inherent in many real-world datasets and in human decision-making, and the Rashomon set makes it explicit (but the unique truth paradigm ignores it).  Essentially, the Rashomon set provides a ``menu'' of good options, from which experts can select models, compare them, or audit.

However, finding the Rashomon set in neural networks is notoriously challenging. While Rashomon sets have been successfully constructed for simpler hypothesis spaces like sparse decision trees \citep{xin2022exploring} or generalized additive models \citep{zhong2023exploring}, these techniques do not scale to neural networks. The continuous, high-dimensional parameter spaces of deep models often admit infinitely many near-optimal solutions that are numerically distinct but \textit{semantically redundant}, meaning they differ in weights but rely on very similar underlying reasoning. Methods such as varying random seeds, applying dropout at inference \citep{Dropout}, or perturbing the final linear layer \citep{donnelly2025rashomon} usually induce diversity concentrated in the output layer or local variation rather than genuine functional divergence.

In this paper, we propose a novel method to explore the Rashomon set of the CBMs by 
constructing and analyzing a Rashomon slice, which provides a representative subset of models with diverse conceptual rationales.
We achieve this through a specialized architecture and a customized objective function. Specifically, we combine small, parallel parameter-efficient adaptation modules with a shared and frozen backbone, enabling the simultaneous learning of multiple models within a single training pipeline. We then apply a customized objective function with a diversity-promoting regularization on the concept representations to force models toward distinct rationales. 
To make joint training scalable, we introduce a model-axis gradient checkpointing scheme that serializes the backward pass across models, reducing memory usage to nearly that of a single model. Together, these components allow us to efficiently find a meaningful set of diverse yet accurate models with minimal additional memory cost.

We evaluate the Rashomon slice quantitatively and qualitatively on multiple datasets (AwA2, CUB, CIFAR-10, CelebA, HAM10000).
Our results show that the slice has rich conceptual diversity without sacrificing accuracy. We demonstrate the practical utility of our method in multiple settings:
\begin{enumerate}[leftmargin=*]
    \item Expert Alignment: Selecting models that prioritize domain-relevant concepts or satisfy trustworthiness criteria (such as fairness) without retraining.

    \item Resolving Inter-Class Confusion: By forcing models toward discriminative rationales, we reduce the accuracy gap in visually similar species (e.g., Indigo Bunting vs. Blue Grosbeak).

    \item Reliable Abstention: We use the disagreement across models as a safety signal to identify ambiguous samples where a single model might otherwise fail.
\end{enumerate}

\textbf{Contributions}: (1) we introduce the first method for exploring the Rashomon set of CBMs; (2) on multiple datasets, our method discovers accurate yet different models, outperforming baselines in efficiency and diversity; (3) we demonstrate three practical use cases, showing that diversity across equally accurate models can be directly leveraged to improve downstream tasks. Our implementation is available at: \url{https://github.com/ShihanF0/CBM-Rashomon-Slice}.




\section{Related Work}
\noindent \textbf{Rashomon Sets.} The Rashomon Effect describes that many different models from a function class can explain a dataset almost equally well \citep{breiman2001statistical}, and the Rashomon set is the collection of these good models \citep{SemenovaRuPa2022, fisher2019all, RudinEtAlAmazing2024}. Existing work in this area can be broadly categorized into efforts to compute and characterize the Rashomon Effect and the Rashomon set \citep{xin2022exploring,zhong2023exploring, hsu2022rashomon, dong2020exploring, molero2026fast} and efforts that study the implications of the Rashomon set for different trustworthy applications \citep{semenova2023path, hsurashomon, boner2024using, ganesh2025curious, black2022model, marx2020predictive, watson2023predictive, turbal2025ellice, hsu2025double}. 

For methods exploring Rashomon sets of neural networks, some vary random seeds during training \citep{d2020underspecification}, use affine transformations to intermediate network activations \citep{eerlings2026diverse}, apply dropout at inference \citep{Dropout}, or perturb the final layer \citep{donnelly2025rashomon}.
Although these approaches produce models with different weights, injected randomness often yields solutions that rely on similar underlying representations, and restricting diversity to the linear layer probes only a small part of the model space. As a result, these methods offer limited insight into alternative reasoning pathways supported by the data, which the Rashomon slice aims to fix.

\noindent \textbf{Concept Bottleneck Models (CBMs).} 
CBMs \citep{koh2020concept} have a special architecture which first predicts concepts from the input and then uses these concepts to predict the label. This two-stage architecture enables users to inspect how concept information is represented and used, and supports human intervention to modify model behavior.
Beyond the standard formulation, recent work explores concept representations by learning richer concept embeddings \citep{espinosa2022concept}, modeling concept uncertainty \citep{ProbCBM}, capturing dependencies between concepts \citep{vandenhirtz2024scbm}, and disentangling concepts in latent space \citep{chen2020concept}. Other extensions formulate inference using energy-based objectives to support concept correction \citep{xu2024ecbm, kim2024eqcbm} or utilize internal ensemble-based gating to refine concept accuracy \citep{ahnEnergyEnsembleConceptBottleneck2025}.
However, these methods focus on a single CBM without considering the Rashomon set. 


\noindent \textbf{Deep Ensembles.} Deep ensembles train multiple networks and aggregate their predictions to improve accuracy, calibration, and out-of-distribution behavior \citep{EnsembleMethods, lakshminarayanan2017deepensembles, Stickland2020DiverseEI, Fort2019DeepEA, yang2020dverge, zhou2018diverse, yang2021trs}. Existing methods encourage members to make different predictions by regularizing for negatively correlated errors among members \citep{Liu1999EnsembleLV,shui2018diversity}, enforcing diversity on saliency maps \citep{dereka2023sdde}, or encouraging hypotheses that fit labeled source data to disagree on unlabeled target data \citep{lee2023diversify}.
To improve efficiency, recent work proposes a framework called LoRA-Ensemble that shares a frozen backbone across members but trains member-specific LoRA updates \citep{muhlematter2024lora}, yielding an architecture that is similar to our Rashomon CBM framework. However, our goal inherently differs from deep ensembles. 
Standard deep ensembles are primarily designed to combine predictions to improve overall accuracy, predictive uncertainty, and out-of-distribution robustness \citep{EnsembleMethods}. 
In contrast, we explicitly encourage differences in concept representations to discover diverse, conceptually distinct ways of solving the task.
Each model stands on its own, which is useful for interpretability and human-in-the-loop design.

\section{Methods}
 \subsection{Preliminaries and Definitions}\label{sec:CBM}
 \paragraph{Concept Bottleneck Models (CBMs).}
Let $\mathcal{D}=\{\x_i, y_i, \bc_i\}_{i=1}^n$ be a dataset, where 
$\x_i \in \mathbb{R}^{s_1 \times s_2 \times s_3}$ is an input image, 
$y_i \in \{1,\dots, K\}$ is the class label with $K$ possible classes, and $\bc_i\in \mathbbm{R}^p$ is a vector of $p$ human-interpretable concepts that describe input $\x_i$.
A typical CBM consists of a concept encoder $g(\cdot): \mathbb{R}^{s_1 \times s_2 \times s_3} \rightarrow \mathbb{R}^p$ that maps input data $\x$ to a latent concept representation $\hat{\bc}=g(\mathbf{x})$, and a classifier $f(\cdot): \mathbb{R}^p \rightarrow \mathbb{R}$ that maps these concepts to output predictions $\hat{y} = f(\hat{\bc})$. 
The empirical risk of a CBM over a dataset $\mathcal{D}$ is defined as the weighted sum of task and concept misclassification losses:
\begin{equation}\label{eq:total}
\mathcal{L}_{\text{total}}(f,g; \mathcal{D}) = \frac{1}{n} \sum_{(\mathbf{x}, y, \mathbf{c}) \in \mathcal{D}} \mathcal{L}_{Y}(f(g(\mathbf{x})), y) + \lambda \mathcal{L}_{C}(g(\mathbf{x}), \mathbf{c}),
\end{equation}
where $\mathcal{L}_{Y}$ and $\mathcal{L}_{C}$ denote the task and concept misclassification losses, respectively, 
and $\lambda$ is a hyperparameter that controls the trade-off between the two losses.

\paragraph{Rashomon Set.} Following \citet{SemenovaRuPa2022}, we define the Rashomon set of CBMs as follows:

\begin{definition}[$\varepsilon$-Rashomon set of CBMs]\label{def:rset}
Given a validation dataset $\mathcal{D}_{val}$, hypothesis spaces $\mathcal{F}$ and $\mathcal{G}$, and a reference model $(f_{\text{ref}}, g_{\text{ref}}) \in \mathcal{F} \times \mathcal{G}$, the $\varepsilon$-Rashomon set $R_{\text{set}}(\varepsilon, \mathcal{D}_{\text{val}})$ is the set of all models $(f, g)$ whose total loss is within a tolerance $\varepsilon$ of the reference loss:
\begin{equation}\label{eq:rset}
R_{\text{set}}(\varepsilon, \mathcal{D}_{val}) \! = \! \left\{ (f, g)\! \in \! \mathcal{F} \!\times\! \mathcal{G} \!:\! \mathcal{L}_{\text{total}}(f, g; \mathcal{D}_{val}) \!\leq\! \mathcal{L}_{\text{total}}(f_{\text{ref}}, g_{\text{ref}}; \mathcal{D}_{val}) \!+ \!\varepsilon \right\}.
\end{equation}
\end{definition}

Typically, the reference model is an empirical risk minimizer $(f_{\textrm{ref}}, g_{\textrm{ref}})\in \arg\min_{f \in \mathcal{F}, g\in \mathcal{G}} \mathcal{L}_{\textrm{total}}(f,g, \mathcal{D})$ \citep{xin2022exploring, SemenovaRuPa2022}. However, obtaining the global minimizer can be challenging for neural networks, and practically, any well-trained CBM can serve as the reference model \citep{fisher2019all}. For example, one approach is to choose as a reference model the one that performs well on the validation data. Similarly, we evaluate the Rashomon set on validation data to mitigate overfitting in neural networks and ensure that the resulting Rashomon set contains models with comparable generalization performance.


\paragraph{Exploring Models in the Rashomon Set.} 
$R_{\text{set}}(\varepsilon, \mathcal{D}_{\text{val}})$ formally contains all models that satisfy the performance bound, but it is typically an uncountable and computationally intractable set to fully compute, particularly when the hypothesis spaces $\mathcal{F}$ and $\mathcal{G}$ are parameterized by neural networks. 
Moreover, even if we could compute many near-optimal neural-network solutions, the resulting collection may still be uninformative: neural networks often admit numerous parameterizations that induce very similar functions, resulting in decision boundaries that are nearly indistinguishable on the data distribution.

Therefore, a more practical alternative is to focus on a finite \emph{diverse} subset of near-optimal models. Unlike previous works that primarily aim to obtain arbitrary members of the Rashomon set \citep{Dropout}, our goal is to identify a set of $M$ models that are both \textit{accurate and meaningfully distinct} in their decision-making logic. Concretely, given a user-defined similarity metric, we seek a set of near-optimal models that are well-separated under this metric.
To do so, we compute a set of $M$ accurate models that minimize total pairwise similarity among its elements. We refer to this set as a \textit{Rashomon slice} and discuss next how we compute it. Since the slice explicitly encourages pairwise separation, it can be viewed as an \emph{approximate} $M$-model packing \citep{SemenovaRuPa2022} of the Rashomon set under the similarity metric.
\subsection{Finding the Rashomon Slice}\label{sec:find-rslice}

To compute the Rashomon slice, we jointly optimize a 
collection of $M$ models using a
single objective that balances predictive performance with an explicit
separation regularizer, encouraging the learned models to remain near-optimal
while promoting diversity across the models.
The objective is optimized on the training data, while Rashomon set membership
(i.e., the performance bound) is evaluated on the validation set $\mathcal D_{\text{val}}$
to better reflect generalization.


\paragraph{Diversity-Aware Minimax Optimization.}
We discover the Rashomon slice by training $M$ models simultaneously. To encourage that every individual model in the slice satisfies the Rashomon performance bound,  
(i.e., $\mathcal{L}_{\text{total}}(f_{\text{ref}}, g_{\text{ref}}; \mathcal{D}_{val}) + \varepsilon$), 
we minimize the maximum task and concept prediction losses across the models. 
As for similarity constraint, 
the primary source of diversity in CBMs lies in their internal concept representations. For two CBMs to represent distinct decision-making strategies, they should produce different concept activation patterns for the same input. Therefore, we maximize diversity by penalizing the average similarity between the predicted concept vectors across models. The resulting joint objective is:
\begin{equation}\label{eq:total_loss}
\min_{f_m, g_m \forall m} \!\left[\max_m \mathcal{L}_{Y}\!(f_m(g_m(\x)), y) \!+\! \lambda \!\cdot\! \max_m \mathcal{L}_{C}(g_m(\x), \bc)
\!-\! \frac{\alpha}{M}\!\sum_{m=1}^M\! \mathcal{L}_{\text{div}}( g_m)\right]\!,
\end{equation}
where $\alpha > 0$ controls the strength of the diversity regularization and the diversity loss for model $m$ is defined as $\mathcal{L}_{\text{div}}(g_m) = 1 - \frac{1}{M-1}\sum_{m'\neq m}\text{sim}(g_m(\x), g_{m'}(\x))$. 
This minimax formulation (first two terms in Eq.~\ref{eq:total_loss}) encourages each model’s total loss to remain small, serving as a proxy to satisfy the validation-based Rashomon bound in Definition \ref{def:rset}. 
$\mathcal{L}_{\text{div}}$ leverages training set cosine similarity as a differentiable surrogate to practically encourage this separation during optimization. 
Overall, this formulation 
encourages the models to remain near-optimal, while the diversity term encourages the models to occupy distinct regions of the decision space.


\paragraph{Parallel Architecture.}
Solving the optimization problem in Eq.~\ref{eq:total_loss} by training $M$ neural networks simultaneously is computationally intensive. To address this, we propose a specialized architecture where all models share a pretrained backbone and utilize model-specific adaptation modules that operate in parallel.

In our framework, each model $m \in \{1, \dots, M\}$ consists of a concept encoder $g_m(\cdot)$ and a classifier $f_m(\cdot)$, i.e., $\hat{y}^{(m)} = f_m(g_m(\mathbf{x}))$. We construct $M$ parallel models by inserting  adaptation modules into specific layers of a shared backbone network $B$ (e.g., ViT). 
Each model $m$ maintains its own set of trainable adaptation modules, denoted as
$A_m = \{A_m^{(1)}, A_m^{(2)}, \dots, A_m^{(L)}\}$, 
where the superscript indicates the layer index at which an module is inserted. 
During a forward pass, only the modules associated with model $m$ are activated, resulting in an effective backbone 
$B_m(\mathbf{x})=B(\mathbf{x}, A_m).$ 
The backbone parameters are shared and frozen across all models, while the adaptation modules $A_m$ are trainable and model-specific (and thus unique). We denote $B^{(l)}$ as shared parameters without these modules of layer $l$.

Given the adapted features $B_m(\mathbf{x})$, each model $m$ uses a set of per-concept linear heads  $\{h_{m,j}\}_{j=1}^p$ to predict the activation of concept $j$. The resulting concept prediction vector is then 
$$    \hat{\bc}^{(m)}=g_m(\x)
    =[h_{m,1}(B_m(\mathbf{x})),
    ...,h_{m,p}
    (B_m(\mathbf{x}))].
$$
This design follows \citet{espinosa2022concept} and \citet{ProbCBM} to improve the flexibility of learning different concept representations.

\begin{figure*}[t]
    \centering
    \begin{subfigure}[b]{0.45\linewidth}
    \includegraphics[width=0.95\linewidth]{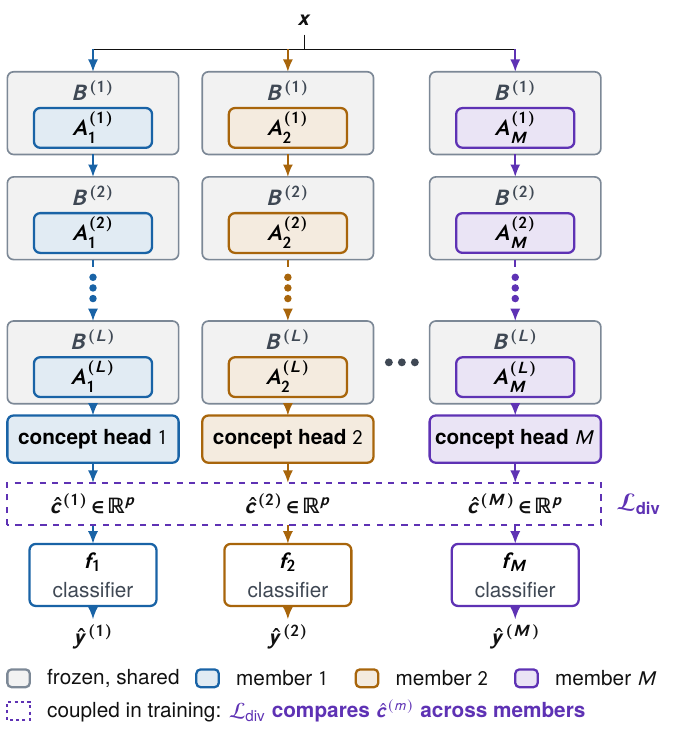}
    \caption{Overall structure}
    \label{fig:overall_structure}
    \end{subfigure}
    \begin{subfigure}[b]{0.45\linewidth}
\includegraphics[width=0.95\linewidth]{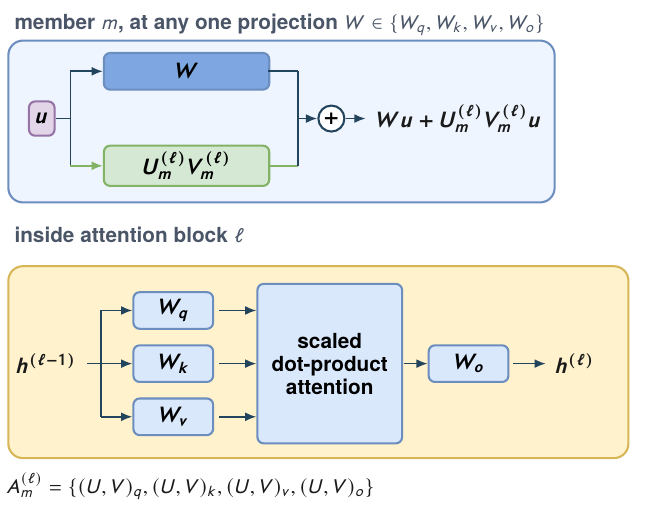}
        \caption{Zoom-in on a ViT attention block}
        \label{fig:lora}
    \end{subfigure}
    \caption{Proposed architecture for finding the Rashomon slice. (a) Overall structure: an input image passes through a shared frozen backbone with member-specific adaptation modules, followed by member-specific concept predictors and classifiers. (b) Zoom-in on a ViT attention block with LoRA: each member applies its own low-rank updates to the Q, K, V, and projection matrices. All LoRA parameters are trained jointly, while at inference each model activates only its own LoRA parameters and shares the frozen backbone weights. }
    \label{fig:architecture}
\end{figure*}

Each classifier $f_m(\cdot)$ then maps the predicted concept vector $\hat{\bc}^{(m)}$ to task label, $\hat{y}^{(m)}=f_m(\hat{\bc}^{(m)})$. Together, the collection $\{g_m, f_m\}_{m=1}^M$ forms a data-driven Rashomon slice. Figure~\ref{fig:overall_structure} summarizes the proposed architecture. 

\noindent\textbf{Implementation.}
Though the proposed architecture is agnostic to the choice of backbone and adaptation mechanism, 
we implement it using a Vision Transformer (ViT) backbone with Low-Rank Adaptation (LoRA) \citep{hulora} as an example. 

\paragraph{Low-Rank Parameterization.} 
Each model $m$ in the Rashomon slice is equipped with a unique set of LoRA modules inserted into the query, key, value, and projection matrices of every attention block. 
For a pretrained weight matrix $W \in \mathbb{R}^{d_{out} \times d_{in}}$, the weight update is constrained to a low-rank decomposition:
$$
W + \Delta W_m = W + U_m V_m
$$
where $U_m \in \mathbb{R}^{d_{out} \times r}$ and $V_m \in \mathbb{R}^{r \times d_{in}}$ are trainable parameters with rank $r \ll \min(d_{in}, d_{out})$. 
The collection of these low-rank parameters across all $L$ blocks, denoted as $\{U_m^{(l)}, V_m^{(l)}\}_{l=1}^L$, constitutes 
$A_m$ that differentiates model $m$ from other members of the Rashomon slice (see Figure \ref{fig:lora}).

\paragraph{Surrogate Losses and Similarity.} 
In the definition of the Rashomon set (Definition \ref{def:rset}), $\mathcal{L}_Y$ and $\mathcal{L}_C$ in Eq. \ref{eq:total} are the task and concept misclassification losses, respectively. However, since these losses, and consequently the first two terms in Eq. \ref{eq:total_loss}, are non-differentiable and cannot be optimized via gradient descent, we use cross-entropy losses as differentiable surrogates during training to optimize Eq. \ref{eq:total_loss}. Specifically, we use multiclass cross-entropy for $\mathcal{L}_Y$ and binary cross-entropy averaged across concepts for $\mathcal{L}_C$.
For the similarity metric $\text{sim}(\cdot,\cdot)$ used in $\mathcal{L}_{\text{div}}$, we compute the cosine similarity between the predicted concept vectors $g_m(\mathbf{x})$ and $g_{m'}(\mathbf{x})$, thereby encouraging models to learn distinct concept representations.

\paragraph{Model-axis Checkpointing Scheme for Efficient Memory Usage.}
Jointly training $M$ CBMs to explore the Rashomon slice is memory-intensive: activation memory scales nearly linearly with $M$, even when conventional layer-wise checkpointing is applied to each model. To avoid this blow-up, we introduce a \textit{model-axis checkpointing scheme}: we wrap each model's full forward (and backward) invocation inside a checkpoint boundary, so that during backpropagation only one model is active and holds its activations at any given time. 
Empirically, 
this reduces peak memory to a level comparable to training a single model, rather than growing linearly with the  size of the slice, which is useful when $M$ is large but GPU memory is limited.

\section{Experiments}

Our evaluation addresses the following questions: (1) how the Rashomon slice compares to baseline methods in discovering diverse yet well-performing models and what qualitative differences arise in how individual models rely on concepts (Section \ref{sec:exp_main}); (2) how the diversity across members can be leveraged for practical downstream tasks (Section~\ref{subsec:usecase}); (3) whether our method remains memory- and parameter-efficient (Section~\ref{sec:efficiency}); and (4) whether the framework generalizes beyond the default ViT+LoRA configuration (Section~\ref{sec:generalizability}).


\paragraph{Datasets.}
We conduct experiments on five datasets: AwA2 \citep{AwA2}, CIFAR-10 \citep{cifar10}, CUB-200-2011 \citep{WahCUB_200_2011}, HAM10000 \citep{tschandl2018ham10000}, and CelebA \citep{celeba}. 
AwA2, CUB and CelebA have human-annotated concepts. Therefore, we directly use them for training. CIFAR-10 and HAM10000 lack explicit concept annotations, and we follow the automatic concept discovery pipeline proposed in \citet{oikarinen2023labelfreeCB} and \citet{truong2024adacbm}, respectively, to get concepts. 

\paragraph{Baselines.}
Since, as discussed in the related work section, we know of no previous method that constructs the Rashomon set for CBMs, we adapt and extend three related approaches to serve as baselines, covering brute-force retraining, inference-time approximation, and diversity-regularized ensembles:
\begin{itemize}[leftmargin=*]
    \item \textit{Random initialization} represents a brute-force method to find a collection of models. We train $M$ CBMs using different random seeds for parameter initialization. Each CBM is fully tunable and updated during training. 
    \item \textit{Dropout CBMs} approximates a Rashomon set through inference-time sampling rather than full retraining. Following \citet{Dropout}, we insert dropout layers throughout the backbone and train a single CBM with dropout enabled. By keeping dropout active during inference and varying the random seed for each pass, we create a set of distinct models, each defined by a unique dropout mask applied to the same trained architecture.
 \item \textit{Diverse Ensemble (DivEnsemble)} is adapted from the diversified deep ensemble method \citep{DiverseNNEnsemble}. 
 It uses a shared, fully tunable backbone $B$ and a shared concept predictor $g$, while each model maintains its own task classifier $f_m$. All models are trained jointly, and the diversity regularization term is applied at the task prediction level.
\end{itemize}

\paragraph{Diversity metrics.}
We evaluate all methods using both concept and task prediction accuracy. Beyond accuracy, we report a set of diversity metrics that provide a more comprehensive analysis of how models differ. 
We consider the following diversity metrics: 
\begin{itemize}[leftmargin=*]
\item \textit{Centered Kernel Alignment (CKA)} compares the concept representations of two models \citep{kornblith2019similarity}. 
Using a linear kernel \(k(\mathbf{x},\mathbf{y}) = \mathbf{x}^\top \mathbf{y}\), we form Gram matrices \(K^{(m_1)}\) and \(K^{(m_2)}\) from the two models’ concept representations evaluated on the same inputs, center them as $\tilde K^{(m)} = H K^{(m)} H$ 
with \(H = I - \tfrac{1}{n}\mathbf{1}\mathbf{1}^\top\). CKA is then computed as the cosine similarity between these centered Gram matrices, normalized by their Frobenius norms. 

\item \textit{SHAP similarity} \citep{lundberg2017unified} measures how similarly different models use learned concepts for prediction. 
We obtain SHAP importance scores over concepts for each model, forming a SHAP importance vector. Diversity is measured by the 1-cosine similarity between the SHAP vectors of two models, where lower similarity indicates more distinct usage of concepts for decision-making. 

\item \textit{Concept coverage} measures the size of the union of the most important concepts across models. For each model, we rank concepts by their SHAP importance scores on the full dataset and select the top 10. We then compute the size of the union of these top-10 concept sets across models. This quantity reflects how much the models’ most influential concepts overlap: a larger union indicates more diverse concept usage.
\end{itemize}
For both CKA and SHAP similarity, we compute all pairwise scores across models to form an \(M \times M\) similarity matrix \(S\) on a fixed test set. We summarize similarity using the off-diagonal mean \(\bar{S}_{\mathrm{off}} = \tfrac{2}{M(M-1)} \sum_{m_1<m_2} S_{m_1m_2}\), where a higher value indicates stronger agreement among models.

\paragraph{Experimental Setup.} We set $M=10$ for each method and use a pretrained ViT-S/16 as the backbone, except for the HAM10000 dataset where we use medical ViT-B/16 \citep{eslami2021does}. The concept predictor $\{h_{m,j}\}_{j=1}^p$ and label classifier $f_m$ are implemented as linear layers. 
To balance task and concept losses, we fix $\lambda=1$, ensuring a fair and standardized comparison across all baselines. However, in practical applications where benchmarking is not the primary goal, one could employ dynamic weighting methods \citep{kendall2018multi} or gradient surgery \citep{yu2020gradient} to optimize this trade-off. During training, we use multiclass cross-entropy for the task loss $\mathcal{L}_Y$ and binary cross-entropy averaged across concepts for the concept loss $\mathcal{L}_C$, as described in Section~\ref{sec:find-rslice}. In our method, we insert LoRA modules into the $Q$, $K$, $V$, and projection matrices of every attention block, with a $0.1$ dropout ratio, a scaling multiplier $s=2$, and a rank of $r=8$ for AwA2, CIFAR-10, and CelebA, $r=16$ for HAM10000, and $r=32$ for CUB. All models are optimized using AdamW, with learning rate and weight decay tuned on a single L40 GPU. 

Once all methods are trained, we first identify the models belonging to the Rashomon set. We select the model from the random initialization baseline that achieves the lowest validation misclassification error-based total loss (see Eq.~\ref{eq:rset}) to serve as a reference model. We set $\varepsilon = 0.03$, corresponding to an absolute performance tolerance of approximately 1.5\% in both task and concept accuracy -- a threshold small enough to ensure all models remain competitive with the reference.
We identify all models whose validation misclassification-based total loss falls within an $\varepsilon$-margin of the reference model's validation misclassification error-based total loss as members of the Rashomon set. We then evaluate the diversity of these members using the following metrics: cosine similarity and centered kernel alignment (CKA) \citep{kornblith2019similarity} of concept predictions, cosine similarity of SHAP importance for the concepts \citep{lundberg2017unified}, and concept coverage (the size of the union of the top-10 most important concepts on the whole dataset evaluated by SHAP). 
More details about datasets and configurations are in Appendix \ref{appendix:exp_setup}.

\subsection{The Rashomon Slice Contains Accurate and Diverse Models}\label{sec:exp_main}

\begin{figure}[t]
     \centering
     \includegraphics[width=\textwidth]{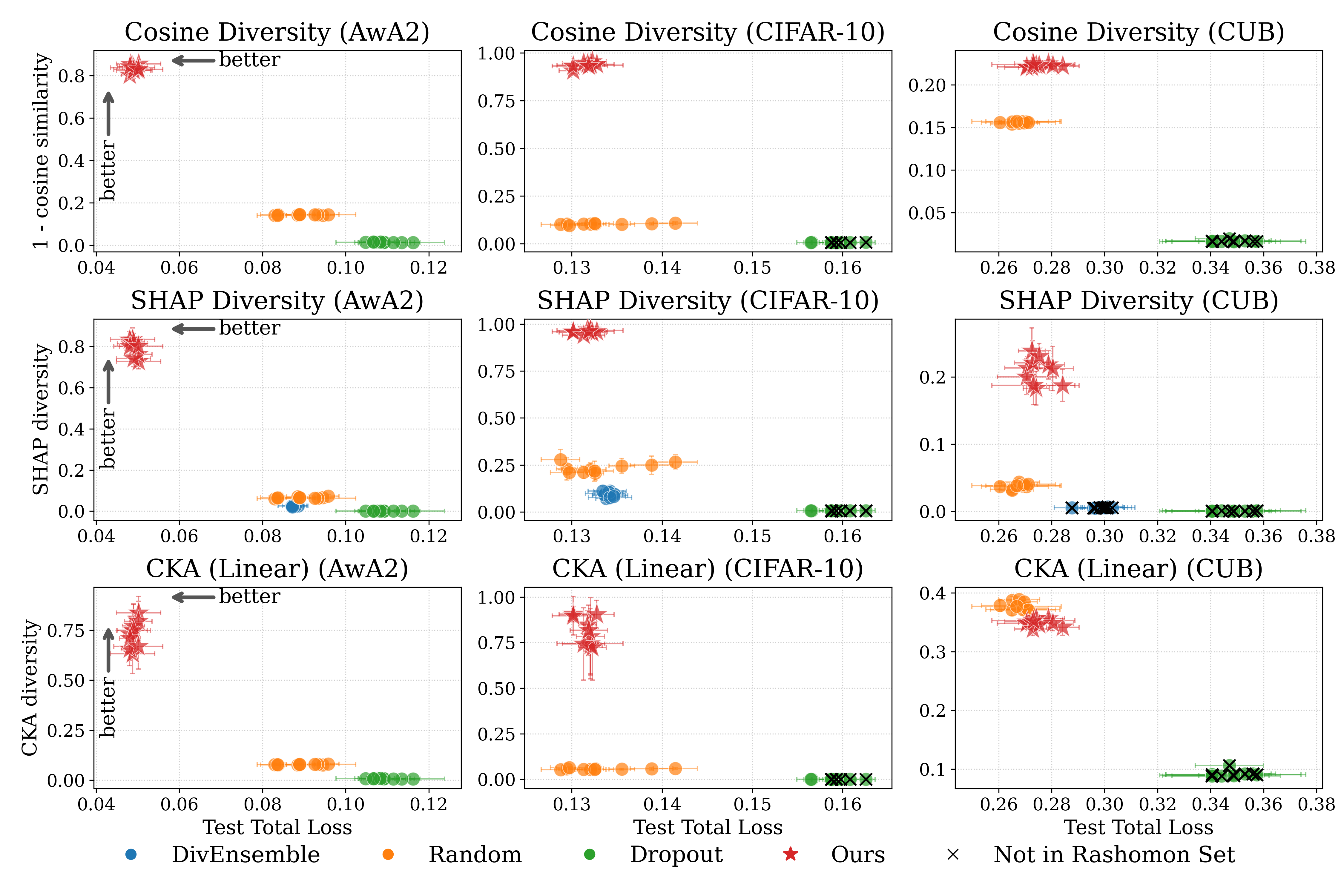} 
     \caption{Comparison of model diversity versus total loss(misclassification loss) on test set across different methods on the AwA2, CIFAR-10, and CUB datasets. The top row displays cosine diversity, the middle row displays SHAP diversity, and the bottom row shows CKA diversity. Each point represents an individual model. Black crosses indicate models that do not satisfy the Rashomon set belonging criterion on the validation data.
     Our method achieves higher diversity while remaining within the Rashomon set.}
     \label{fig:task_loss_vs_div}
\end{figure}

\paragraph{Quantitative Results.} \label{subsec:rashomon_diversity}
In this section, we show that our method can find more diverse models yet stay within the Rashomon set (See Figure \ref{fig:task_loss_vs_div}). It is important to note that each point in Figure \ref{fig:task_loss_vs_div} represents an individual model. We evaluate the standalone performance of each model obtained from different methods.

As we can see, our method successfully finds models (red stars) that consistently achieve low test total loss with no crosses on them, indicating that all of these models fall within the $\varepsilon-$threshold of the reference model.  
Meanwhile, models found by our method also stay in the top-left region across the AwA2, CIFAR-10, and CUB datasets. This indicates that our method is able to find models that achieve both low test total loss and higher diversity measured by cosine similarity, SHAP, and CKA compared to baselines. 
Beyond pairwise diversity, we report the concept coverage -- the size of the union of the top 10 most important concepts based on SHAP across all models from a given method that successfully fall within the Rashomon set. Table \ref{tab:combined_side_by_side} shows that our method achieves the highest coverage, indicating that it finds models that rely on a broader set of concepts for prediction. 
It is worth noting that both our method and some baselines achieve a concept coverage of six on the CelebA dataset. This is because our setup of CelebA provides only six available concepts, which limits the space of concept-based explanations.
More results are shown in Appendix \ref{appendix:mainexp}. 



\definecolor{lightgray}{gray}{0.93}
\begin{table}[t]
    \centering
    \caption{Comparison of concept coverage across five datasets for models \textbf{in the Rashomon set}. N/A indicates that no models produced by that method fell within the Rashomon set.} 
    \label{tab:combined_side_by_side}
     \small
        \centering
        \label{tab:concept_coverage}
        \begin{tabular}{@{} l ccccc @{}}
            \toprule
            \textbf{Method} & \textbf{AwA2} & \textbf{CIFAR10} & \textbf{CUB} & \textbf{HAM10000} & \textbf{CelebA}\\
            \midrule
            DivEnsemble  & 33 & 19 & N/A & N/A & N/A \\
            Random   & 43 & 36 & 51 & 36 & \textbf{6} \\
            Dropout  & 19 & 14 & N/A & N/A & \textbf{6}\\
            \textbf{Ours} & \textbf{59} & \textbf{61} & \textbf{63} & \textbf{57} & \textbf{6} \\
            \bottomrule
        \end{tabular}
\end{table}

\begin{figure}
    \centering
    \includegraphics[width=0.7\linewidth]{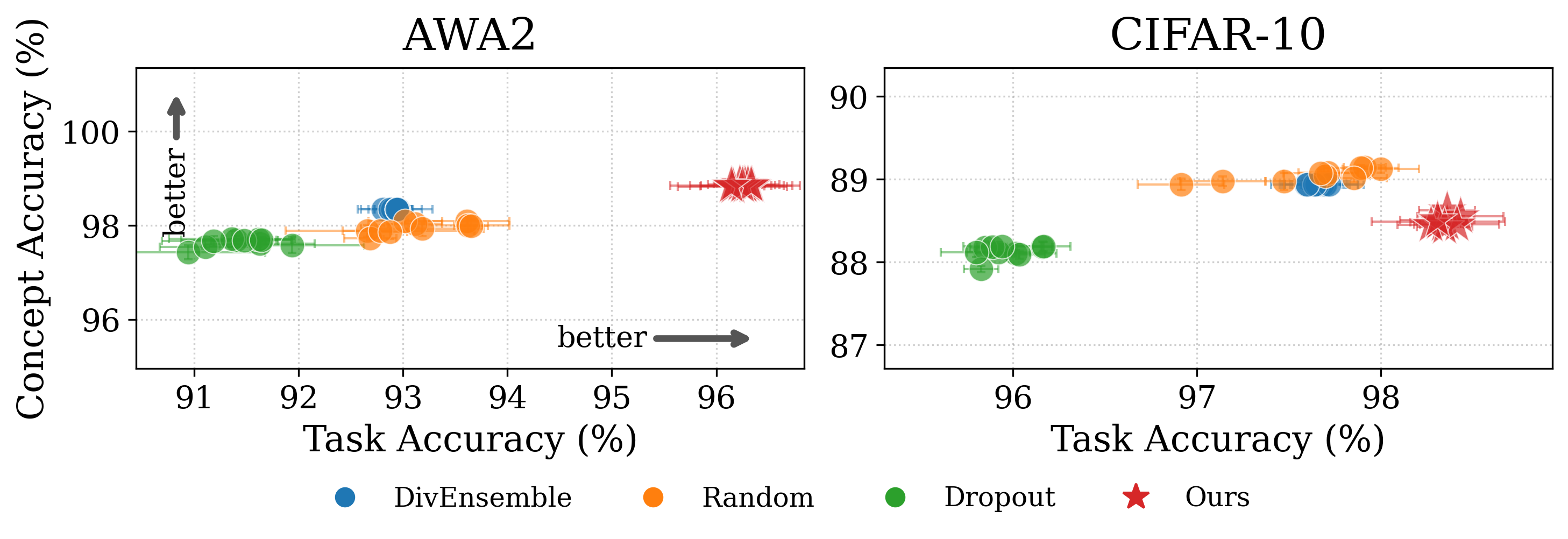}
    \caption{Per-model task and concept accuracy on AwA2 and CIFAR-10. Each point represents an individual model.  Our method achieves task accuracy comparable to or better than baselines, with stable variance across members and only a marginal decrease in concept accuracy on the CIFAR-10 dataset.}
    \label{fig:acc_comparison}
\end{figure}

Figure~\ref{fig:acc_comparison} shows that on both AwA2 and CIFAR-10 datasets, the models found by our method attain task accuracy that is comparable to or better than the baselines with low variance across Rashomon members. Concept accuracy decreases marginally compared to the
strongest baselines. Combining with Figure \ref{fig:task_loss_vs_div} and Figure \ref{fig:result_ham_celeba}, the results show that our method achieves high concept accuracy and high cosine diversity simultaneously, so there is enough inherent data diversity for the cosine penalty to not fight the concept loss for reasonable diversity regularization. Instead, it selects among the multiple high-accuracy concept encoders that the data already admits.


\begin{figure}[t]
    \centering
    \includegraphics[width=\linewidth]{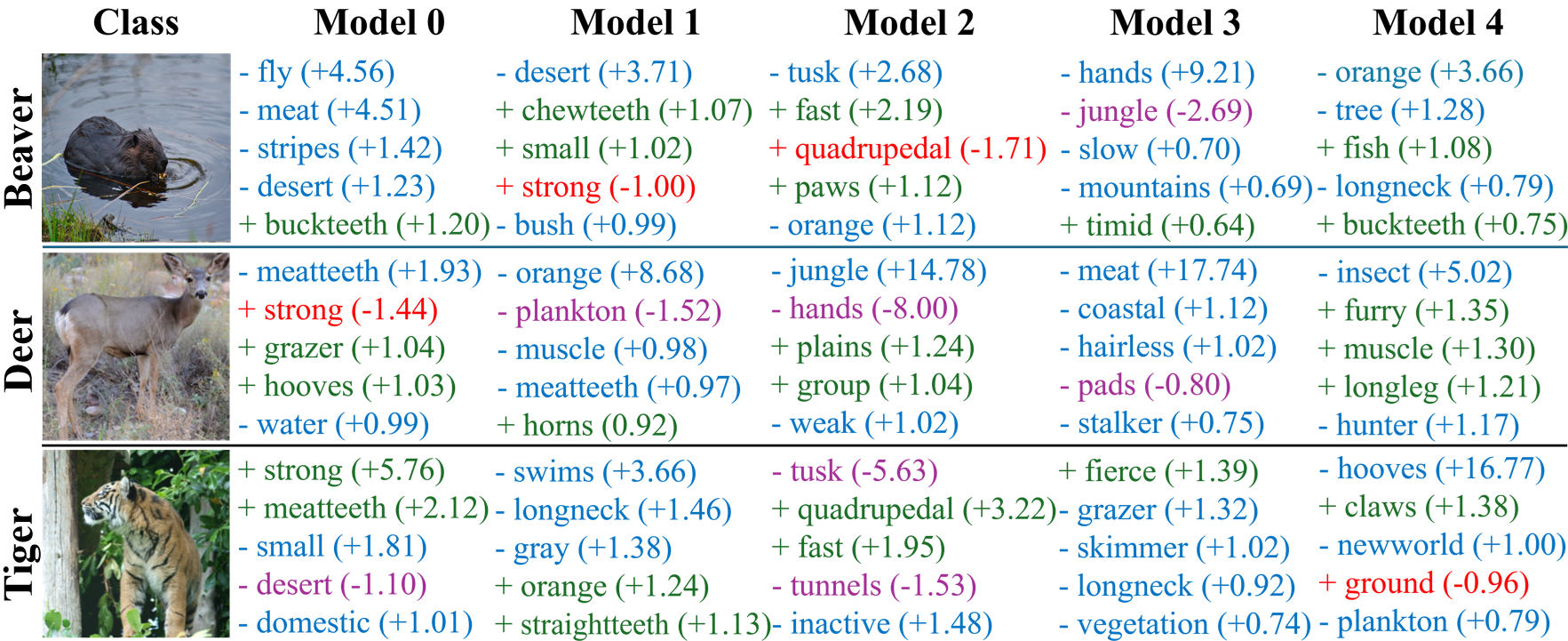}
    \caption{Top-5 concepts based on contribution scores $W_{p,k}\hat{c}_p$ of each concept $p$ to the prediction of class $k$ (value in parentheses) for beaver (top), deer (middle), and tiger (bottom) classes across five Rashomon slice members. The sign stands for whether the concept is predicted to be present ($+$, $\hat{c}_p > 0$) or absent in the corresponding class ($-$, $\hat{c}_p \leq 0$). All the predictions for concepts shown are correct and correspond to ground-truth concept labelings for the classes. Colors indicates the decision rationale in which the concept contributes to the prediction: \textcolor{teal}{green denotes positive evidence ($W_{p,k}>0, \hat{c}_p>0$);} \textcolor{red}{red represents negative evidence ($W_{p,k}\leq 0, \hat{c}_p>0$);} \textcolor{cyan}{blue means excluding evidence ($W_{p,k}>0, \hat{c}_p\leq0$);} \textcolor{violet}{purple stands for spurious evidence ($W_{p,k}\leq 0, \hat{c}_p\leq 0$).}
    The diverse patterns highlight that different members rely on different rationales.}
    \label{fig:qual_anal_AwA2}
\end{figure}

\paragraph{Qualitative Results.} \label{sec:qualitative_5} 
In this section, we provide qualitative results for the diversity of models discovered by our method. We visualize five arbitrary models from the Rashomon slice for the prediction of \textbf{beaver}, \textbf{deer}, and \textbf{tiger} classes, highlighting the top five concepts that each model relies on. As shown in Figure~\ref{fig:qual_anal_AwA2}, different  members of the slice justify the same prediction using different rationales. 
We categorize these concepts into four decision rationale groups: positive evidence (having it supports the prediction, green), negative evidence (having it has a negative effect on the predictions, red), excluding evidence (does not having it supports the predictions, blue), and spurious correlation (the concept is problematic, purple). 

\begin{figure}[t]
    \centering
    \includegraphics[width=\linewidth]{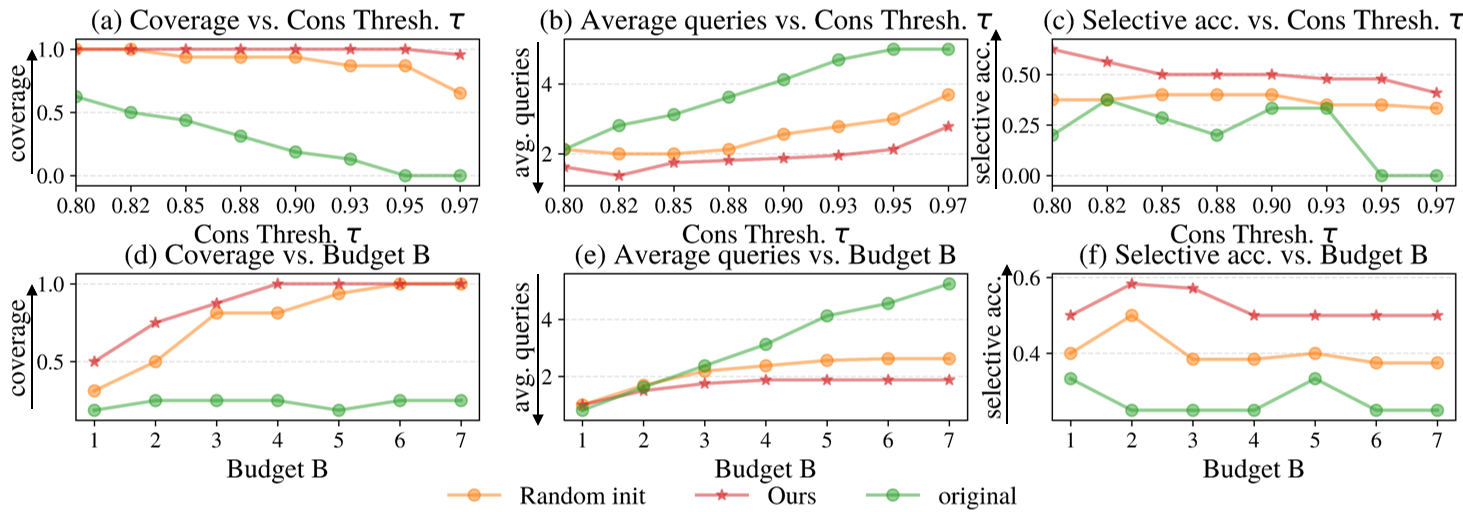}
    \caption{Performance of abstention on Dermatofibroma class from the HAM10000 dataset across varying the confidence thresholds $\tau$ (top row) and query budgets $B$ (bottom row) on the coverage (fraction of non-abstained predictions), average number of concept queries, and selective accuracy (accuracy on non-abstained samples). Arrows indicate the preferred direction. Compared to the original single-CBM safeguard (\textit{original}) and the random initialization baseline, the Rashomon-member safeguard achieves higher coverage with fewer queries while improving selective accuracy across settings.}
    \label{fig:safeguard_HAM10000_class3}
\end{figure}

Taking \textbf{beaver} class as an example, the five Rashomon members justify the same prediction through different conceptual strategies. Model 0 and Model 4 rely almost exclusively on exclusion reasoning, identifying absent traits from different perspectives such as ``fly'', ``meat'', ``stripes'', and ``desert'' for Model 0 and ``orange'', ``tree'' and ``longneck'' for Model 4 to rule out non-beaver classes, while using a small set of  physical characteristics as direct evidence. 
Model 1, on the other hand, combines positive evidence from physical traits (``chewteeth'', ``small'') with environmental exclusion (``desert'', ``bush'') and treats ``strong'' as negative evidence. Model 2 also mixes positive physical traits (``fast'', ``paws'') with excluding evidence (``tusk'', ``orange''), while using ``quadrupedal'' as negative evidence. Model 3 is dominated by a single large exclusion signal ``hands'' and is complemented by two behavioral traits ``timid'' and ``slow'' as positive and excluding evidence respectively. Interestingly, Model 3 also includes a spurious concept ``jungle'', which means that the model has a wrong perception of the surroundings. Overall, the five members span different combinations of decision-making rationales and use different profiles of traits ranging from physical, behavioral to contextual, illustrating that equivalent predictions within a Rashomon set can arise from fundamentally different concept-level decision mechanisms.

Beyond inspection, the Rashomon slice enables personalized intervention by allowing a practitioner to select a member whose concept reliance best matches domain preferences. For example, if a user finds a highly weighted concept, such as \textit{jungle} in Model~2 for the \textbf{deer} class, undesirable or potentially spurious, they can instead deploy a different Rashomon slice member that achieves similar accuracy while placing less emphasis on that concept (e.g., Model~4). This provides a lightweight alternative to retraining: practitioners can choose among models in the slice based on the concepts they want to encourage or avoid. We discuss more about  model choice and other use-cases in Section \ref{subsec:usecase}.

\paragraph{Diversity Analysis.} 
A diverse Rashomon slice is useful in practice if its members genuinely reason differently.
Just like DivEnsemble, if members just re-weight the same underlying concept representations through different output heads, changing between models can reduce a concept's influence on the final prediction, but cannot offer a genuinely different way of perceiving the input. In other words, since the underlying concept detectors remain identical, any spurious or flawed concept detection will happen across all members.
Since our architecture gives each member its own LoRA modules inside the shared backbone, we expect diversity to emerge at the level of concept representations themselves, not just at the output head. To show this, we performed intervention experiments on the tiger class on the AwA2 dataset, and also traced the origin of concept-representation diversity through layer-wise eigenvector visualizations with details in Appendix~\ref{app:source_of_diversity}. Overall, we observed that diversity is concentrated in the deeper LoRA modules rather than the shallow layers and confirmed that representational diversity in the Rashomon slice models is architecturally grounded.

\subsection{Use Cases: The Rashomon Slice Can Help with Abstention, Mitigation, and Trustworthy Model Selection}
\label{subsec:usecase}
In this section, we provide three distinct use cases once we have the Rashomon slice for CBMs.

\paragraph{1. Efficient and Reliable Abstention.}

In cases when CBM is not confident, prior work introduced conceptual safeguards \citep{joren2024classification} that abstain from predicting on high-uncertainty samples and instead seek human intervention to confirm key concepts. Traditional safeguards rely on Monte Carlo sampling from a single model to quantify uncertainty. However, this often reflects local variation around a single rationale. A Rashomon set provides a more natural framework for this task, as it contains multiple, near-optimal, but conceptually distinct rationales for the same input. Therefore, we use individual Rashomon members to estimate uncertainty. If the members disagree on a task prediction beyond a threshold $\tau$, the system abstains from the sample and queries an expert for a specific concept label (within a budget $B$). We then reweight the members based on their consistency with the expert’s feedback, repeating this process until the weighted consensus exceeds $\tau$ or the budget $B$ is exhausted.

We compare our Rashomon-member safeguard against the single-CBM safeguard and the random initialization baseline across various budgets $B$ and thresholds $\tau$ on both Beaver class from the AwA2 dataset (see Appendix \ref{appendix:usecase}) and Dermatofibroma class from the HAM10000 dataset. As shown in Figure \ref{fig:safeguard_HAM10000_class3}, our method returns a prediction on a larger fraction of test examples without needing to abstain, even at high confidence thresholds (left column). Our models require fewer human concept queries to reach a confident consensus (middle column), and for the samples where a prediction is provided, the accuracy is higher than the baselines (right column). These results indicate that diverse Rashomon members can provide a more precise indicator of uncertainty than sampling from a single model. 

\paragraph{2. Trustworthy Model Selection.}

\begin{wrapfigure}[21]{r}{0.42\columnwidth}
    \centering
    \includegraphics[width=0.4\columnwidth]{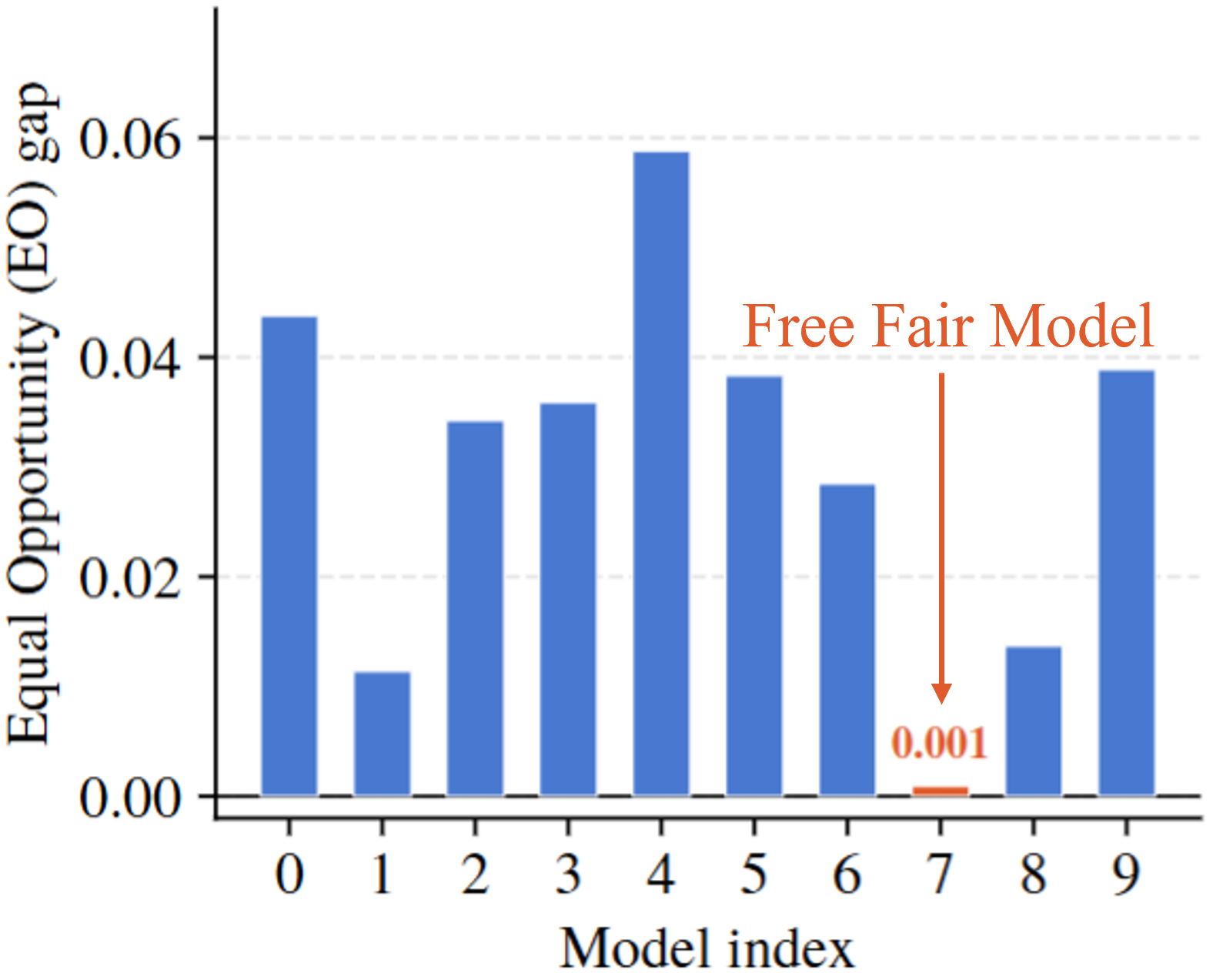}
    \caption{The Rashomon slice finds a fair model for free (Young $\to$ Wavy Hair). All ten models achieve comparable accuracy, but Equal Opportunity (EO) gaps span $[0.001, 0.059]$. Model 7 achieves near-zero EO gap at no accuracy cost.}
    \label{fig:fairness_main}
\end{wrapfigure}


Beyond diversity, the Rashomon slice enables a broader form of trustworthy model selection: since all members achieve near-identical accuracy by construction, any variation in deployment-relevant properties across members is a free gain with no accuracy trade-off required. Whenever a practitioner has a secondary trustworthiness objective in mind, they can simply select the Rashomon member that best satisfies it, rather than retraining under complex constraints \citep{black2022model, hsurashomon}. We demonstrate this principle with fairness using the constructed Rashomon slice.

On CelebA, despite similar accuracy, Equal Opportunity gaps span $[0.001,0.059]$ across Rashomon slice members (Figure~\ref{fig:fairness_main}), indicating that fairness disparity is an artifact of a particular decision rationale rather than an inherent property of the task and a fairer model can be obtained for free, a gain which is unattainable for single-model pipelines.
The same principle might extend naturally to other trustworthiness properties such as robustness or privacy, since each Rashomon slice member emphasizes a distinct set of concepts and may therefore interact differently with distributional shifts or membership inference \citep{hsu2025double}.

\begin{table}[htbp]
\centering
\caption{Conceptual collapse and class-wise accuracy gaps for confounded class pairs. 
Random initialization yields highly aligned concept rationales and large inter-class gaps, whereas Rashomon slice reduce cross-model concept similarity and shrink the accuracy gap.}
\resizebox{\linewidth}{!}{%
\begin{tabular}{lcccc}
\toprule
 & Random Init & Dropout & DivEnsemble & Ours \\
\midrule
\multicolumn{5}{l}{\textbf{Indigo Bunting vs.\ Blue Grosbeak}} \\
\cmidrule{1-1}
Avg Cross-Model Concept Cos Sim $\downarrow$ & 0.8045 & 0.9770 & N/A & \textbf{0.6085} \\
Avg Acc (IB/BG) $\uparrow$ & 88.67\% / 72.67\% & 82.67\% / 58.67\% & 86.67\% / 68.00\% & \textbf{92.00\% / 81.33\%} \\
Avg Acc Gap (IB-BG) $\downarrow$ & 16.00\% & 24.00\% & 18.67\% & \textbf{10.67\%} \\

\cmidrule{1-1}
\multicolumn{5}{l}{\textbf{Bobcat vs.\ Leopard}} \\
\cmidrule{1-1}
Avg Cross-Model Concept Cos Sim $\downarrow$ & 0.8549 & 0.9883 & N/A & \textbf{0.1629} \\
Avg Acc (BC/LP) $\uparrow$ & 94.21\% / 99.58\% & 93.65\% / 98.96\% & 93.10\% / 99.93\% & \textbf{98.17\% / 99.93\%} \\
Avg Acc Gap (LP-BC) $\downarrow$ & 5.37\% & 5.31\% & 6.83\% & \textbf{1.76\%} \\
\bottomrule

\end{tabular}%
}
\label{tab:confounding}
\end{table}

\paragraph{3. Confounding Classes.}
 Fine-grained datasets often contain confounding class pairs with substantial visual overlap (e.g., CUB: Indigo Bunting vs.\ Blue Grosbeak; AwA2: Bobcat vs.\ Leopard), where a model may lock onto evidence that is predictive for one class but insufficiently discriminative for its confounder, resulting in a large per-class accuracy gap. We address this by using the Rashomon slice as a diversified ensemble: rather than relying on a single decision rationale, we aggregate across members whose diverse concept representations collectively cover more discriminative evidence. This mitigates conceptual collapse, where independently trained models converge on the same shared, potentially non-discriminative concepts. As shown in Table~\ref{tab:confounding}, the random initialization baseline exhibits strong conceptual collapse (cross-model cosine similarity $\geq 0.8$) and a large accuracy gap between the two confounded classes. In contrast, Rashomon slice reduces cross-model concept similarity substantially, and as a result both per-class accuracies improve and the gap between them shrinks.

\subsection{Memory and Parameter Efficiency} 
\label{sec:efficiency}
We also report the memory cost and the number of trainable parameters for each method on AwA2 dataset in Table \ref{tab:mem_param_and_ablation} (left). Our method is very memory efficient, using only 7.6\% of the memory and 1.2\% of the trainable parameters required by the method based on random initialization. This substantial reduction stems from LoRA and the model-axis checkpointing schema. Further, ablations in Table \ref{tab:mem_param_and_ablation} (right) show that these design choices 
reduce different components of the memory (parameters vs. activations) and their effects accumulate. Results for ViT+LoRA on more datasets are available in Appendices \ref{app:memory} and \ref{app:mem_ablation}. Additional time efficiency analysis is available in \ref{app:time}.

\definecolor{lightgray}{gray}{0.93}

\begin{table}[htbp]
    \centering
    \small
    \caption{
    \textbf{Left:} Exact and relative memory and trainable-parameter costs on AwA2. Random initialization serves as the baseline for ratios.
    \textbf{Right:} Peak GPU memory under different combinations of LoRA and checkpointing on AwA2. In our method, checkpointing and LoRA are designed to reduce different components of the memory footprint, and combining them gives the largest memory reduction.
    }
    \label{tab:mem_param_and_ablation}

    \begin{subtable}[t]{0.48\textwidth}
        \centering
        \begin{tabular}{@{}lrrr@{}}
            \toprule
            \textbf{Metrics} & \textbf{Rand.} & \textbf{DivE.} & \textbf{Ours} \\
            \midrule
            Mem (GiB) & 37.09 & 4.66 & \textbf{2.81} \\
            \rowcolor{lightgray}
            Mem (\%) & 100.0 & 12.5 & \textbf{7.6} \\
            \# Params (M) & 217.0 & 21.7 & \textbf{2.7} \\
            \rowcolor{lightgray}
            \# Params (\%) & 100.0 & 10.0 & \textbf{1.2} \\
            \bottomrule
        \end{tabular}
    \end{subtable}
    \hfill
    \begin{subtable}[t]{0.48\textwidth}
        \centering
        \begin{tabular}{@{}lrr@{}}
            \toprule
            \textbf{Variant} & \textbf{Mem (GiB)} & \textbf{Reduction (\%)} \\
            \midrule
            no-LoRA / no-ckpt & 37.09 & 0\% \\
            no-LoRA / ckpt    & 4.66  & 87.4\% \\
            LoRA / no-ckpt    & 25.54 & 31.1\% \\
            \textbf{LoRA / ckpt} & \textbf{2.81} & \textbf{92.4\%} \\
            \bottomrule
        \end{tabular}
    \end{subtable}
\end{table}

\subsection{Generalizability beyond ViT+LoRA Architecture}
\label{sec:generalizability}
We further evaluate whether our method beyond ViT + LoRA configuration. 
We add two additional implementation settings:
(1) ViT-S/16 with DoRA, which replaces LoRA with a different adaptation method and
(2) ResNet-18 with ConvAdapters, trained jointly, while at inference each model activates only its own LoRA parameters and shares the frozen backbone weights. In each setting, all baselines (Random, Dropout, DivEnsemble) are also implemented with the same backbone.

\begin{figure}[htbp]
    \centering
    \includegraphics[width=\linewidth]{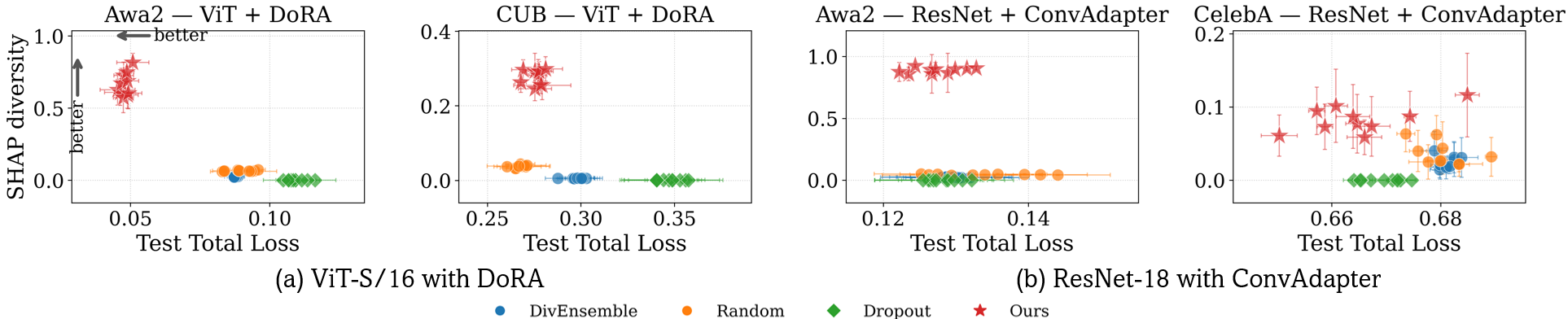}
    \caption{\textbf{Generalizability beyond ViT+LoRA.} We evaluate our method under two additional settings: ViT-S/16 with DoRA and ResNet-18 with ConvAdapters. Each point corresponds to one model in the Rashomon slice. Our method achieves higher SHAP diversity at comparable test total loss, suggesting that the framework generalizes beyond the ViT+LoRA configuration.}
    \label{fig:generalizability}
\end{figure}

As shown in Figure~\ref{fig:generalizability}, in both settings our method identifies Rashomon slices with better or comparable task loss and higher SHAP diversity than baselines using the same backbone. These results indicate that the proposed framework is not tied to certain backbones or adaptation modules, but can be applied to different parameter-efficient adaptation mechanisms and backbone architectures.

\section{Conclusions and Future Work}

This work introduced the first lightweight framework for constructing a diverse set of equally accurate CBMs that are distinct in their concept-level decision rationales from a single training process. We demonstrated that access to this Rashomon slice unlocks practical capabilities unavailable to single-model approaches, including reliable abstention by treating member disagreement as an uncertainty signal, trustworthy model selection by identifying members that satisfy secondary objectives such as fairness at no accuracy cost, and resolution of inter-class confusion by leveraging diverse discriminative concept-level reasoning rationales across members. Taken together, these use cases show that model multiplicity in CBMs is a practical resource that practitioners can actively exploit for downstream tasks.

At the same time, making model multiplicity explicit may create opportunities for post hoc cherry-picking \citep{black2022model}. There are multiple steps one can take to mitigate this risk, including prespecifying selection criteria applied uniformly on held-out data, auditing members for spurious-concept reliance, or reporting the full slice alongside any selected member.
Several directions also remain open for future work. The number of models $M$ is currently user-specified. Developing data-driven criteria to determine it automatically remains an important extension. Our method requires more training epochs to converge than single-model 
baselines, a common trade-off with parameter-efficient adaptation, which we view as an acceptable cost given the substantial memory savings that allow many more slice members to be trained under fixed GPU budgets. Nonetheless, developing training schedules that narrow this gap remains a future direction.
Introducing sparsity regularization could encourage each member to rely on a smaller, more interpretable subset of concepts. Finally, interactive visualization tools that allow domain experts to browse and compare Rashomon members would make the diversity of concept-based reasoning strategies accessible beyond machine learning practitioners. 

%
%
\bibliographystyle{tmlr}
\bibliography{refs}

\clearpage
\appendix
\renewcommand{\theHsection}{\thesection.appendix}

\section{Experimental Setup}\label{appendix:exp_setup}

\subsection{Datasets}
\label{appendix:data}
In our experiments, we evaluate the proposed method across five datasets that span both fine-grained and coarse-grained visual domains. This allows for a comprehensive assessment of our model's performance under different concept supervision settings, including both human-annotated concepts and automatically discovered concepts.
\paragraph{Animals with Attributes 2 (AwA2) \citep{AwA2}.}The AwA2 dataset is a widely used benchmark for zero-shot learning and concept-based models. It consists of 37,322 images spanning 50 animal classes. The dataset provides 85 human-annotated binary attributes (e.g., ``stripes'', ``water'', ``domestic''). We directly utilize these expert-defined attributes as our ground-truth concept space for training, following the standard train-test splits provided in the original benchmark.
\paragraph{Caltech-UCSD Birds-200-2011 (CUB) \citep{WahCUB_200_2011}.}The CUB dataset is a fine-grained image classification dataset containing 11,788 images across 200 bird species. It includes dense annotations with 312 binary attributes representing specific bird parts and characteristics (e.g., ``wing color'', ``beak shape''). Similar to AwA2, these human-annotated attributes serve directly as the concept-level supervision in our experiments, providing a high-dimensional and highly granular concept space.
\paragraph{CIFAR-10 \citep{cifar10}.}CIFAR-10 contains 60,000 images across 10 coarse-grained object classes. Because CIFAR-10 lacks explicit concept annotations, we construct the concept space using the automatic, label-free concept discovery pipeline proposed by \citet{oikarinen2023labelfreeCB}. This involves using a Large Language Model to generate a comprehensive candidate set of descriptive concepts for each class, which are then filtered and aligned using a pretrained CLIP model to ensure the concepts accurately reflect the visual semantics of the images.
\paragraph{HAM10000 \citep{tschandl2018ham10000}.} HAM10000 is a medical dataset that consists of 10,015 dermatoscopic images of pigmented skin lesions across 7 diagnostic categories. Since explicit concept labels are unavailable, we construct concepts following the method in  \citet{truong2024adacbm}. Specifically, we start with the class-specific concept pool Chowdhury et al. constructed using GPT-4. To identify a compact and highly discriminative subset of concepts, we apply a selection procedure: for each of the 7 task classes, candidate concepts are ranked using a $t$-test on their CLIP similarity margins computed on the training set. We select the top 20 concepts per class and take the union across all classes, leading to 139 unique binary concepts.
\paragraph{CelebFaces Attributes (CelebA) \citep{celeba}.}CelebA is a large-scale face attributes dataset containing over 200,000 celebrity images. Following the experimental setup in prior concept-based learning research \citep{espinosa2022concept}, we filter the dataset to focus on the eight most balanced facial attributes and use them to construct both concept and task supervision. Specifically, the first six attributes are used as the binary concepts. To ensure a structured target space with sufficient label diversity, we use all eight attributes to define $2^8 = 256$ composite classes.

\subsection{Configurations}
\label{appendix:config}
We use a pretrained ViT-S/16 as the backbone, except for the HAM10000 dataset where we use a medical ViT-B/16 \citep{eslami2021does}. For all methods, the concept predictor $\{h_{m,j}\}_{j=1}^p$ and label classifier $f_m$ are implemented as linear layers. We set $\lambda=1$ to achieve a balance between concept and task accuracy and train the model using the AdamW optimizer with learning rate in $\{0.0001,0.0003,0.00075\}$ and weight decay in $\{0,0.0004\}$ on a single L40 GPU. 

For our method, we insert LoRA modules into the $Q$, $K$, $V$, and projection matrices of every attention block, with scaling multiplier $s=2$, dropout ratio=0.1, and a rank of $r=8$ for AwA2, CIFAR-10 and CelebA dataset. We set $r=16$ and $r=32$ for HAM10000 and CUB dataset respectively to increase the capacity of LoRA modules. All backbones are frozen, and only the adapter parameters and concept heads are updated during training.

Another hyperparameter we have is $\alpha$ (see Eq.~\ref{eq:total_loss}). We dynamically update $\alpha$ during training based on the gradient magnitude.  Specifically, let $g^{(t)} = \mathbb{E}_{w\in\mathcal{P}} \big[|\nabla_{w}\mathcal{L}_{\text{total}}^{(t)}|\big]$ denote the average absolute gradient across all parameters $\mathcal{P}$ in the concept heads. To ensure the regularization strength remains within a valid range $[\alpha_{\text{min}}, \alpha_{\text{max}}]$, we map the gradient using a shifted and scaled sigmoid function $\sigma$:
\[\alpha^{(t+1)} = \alpha_{\text{min}} + 2(\alpha_{\text{max}} - \alpha_{\text{min}}) \left(\sigma(g^{(t)}) - 0.5\right). \] 
$\alpha_{\max}$ and $\alpha_{\min}$ are tunable hyperparameters to further control the strength of diversity regularization. Intuitively, in early stages of training, the actual $\alpha$ will be close to $\alpha_{\max}$ and gradually decay to $\alpha_{\min}$ as the model converges. In our implementation, we tune $\alpha_{\max}$ in \{0.75,1\} and $\alpha_{\min}$ in \{0.25,0.5,0.75\}.

\begin{figure}[htbp]
    \centering
    \includegraphics[width=0.995\linewidth]{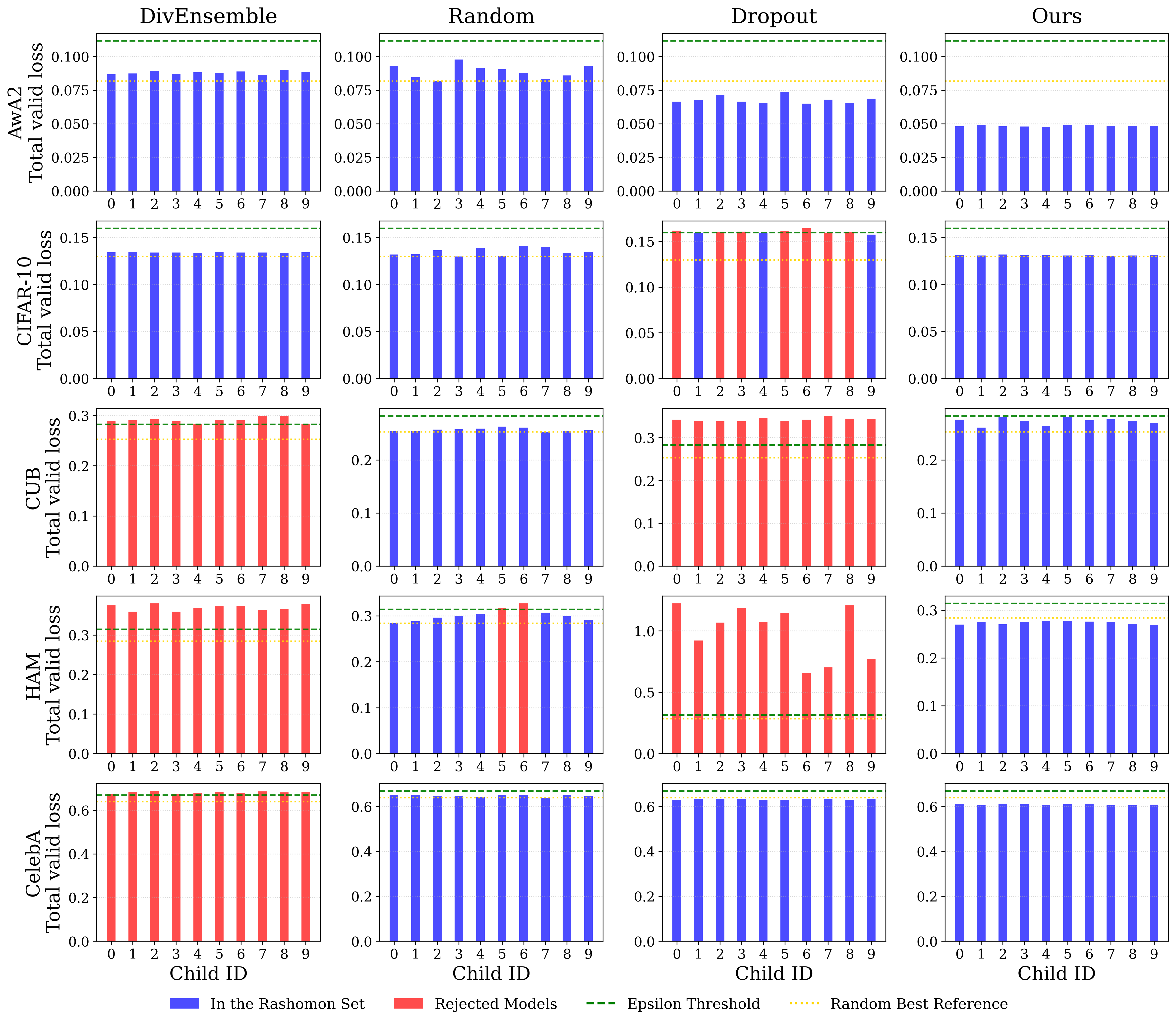}
    \caption{Identification of Rashomon Set Membership across Five Datasets. Membership is determined using the validation misclassification-based total loss defined in Eq.~\ref{eq:total}. Blue bars indicate models within the Rashomon set, red bars indicate rejected models, and the green dashed line denotes the $\varepsilon$-threshold.
    The reference performance (yellow dotted line) is determined by the best model on the validation set from the random initialization baseline.}
    \label{fig:rashomon_inclusion}
\end{figure}

\section{Additional Quantitative Results} \label{appendix:mainexp}
\subsection{Rashomon Set Membership of Rashomon slice and Baselines}
In this section, we evaluate whether the models discovered via different methods successfully stay within the $\varepsilon$-Rashomon set as defined in Definition \ref{def:rset}. 
We select the model from the random initialization baseline that achieves the lowest validation misclassification error-based total loss (see Eq.~\ref{eq:rset}) to serve as a reference model. We set $\varepsilon = 0.03$, corresponding to an absolute performance tolerance of approximately 1.5\% in both task and concept accuracy -- a threshold small enough to ensure all models remain competitive with the reference.
We identify all models whose validation loss falls within an $\varepsilon$-margin of the reference model's validation total loss as members of the Rashomon set. 

Figure \ref{fig:rashomon_inclusion} shows the membership of the Rashomon set across different methods. The green dashed line denotes the Rashomon bound. Blue bars represent models whose validation loss falls within this $\varepsilon$-margin, and red bars indicate models that exceed it. 
As shown in Figure \ref{fig:rashomon_inclusion}, 
models found by our method are usually stay within the Rashomon set across all five datasets.

\begin{figure}[ht]
    \centering
    \includegraphics[width=0.98\linewidth]{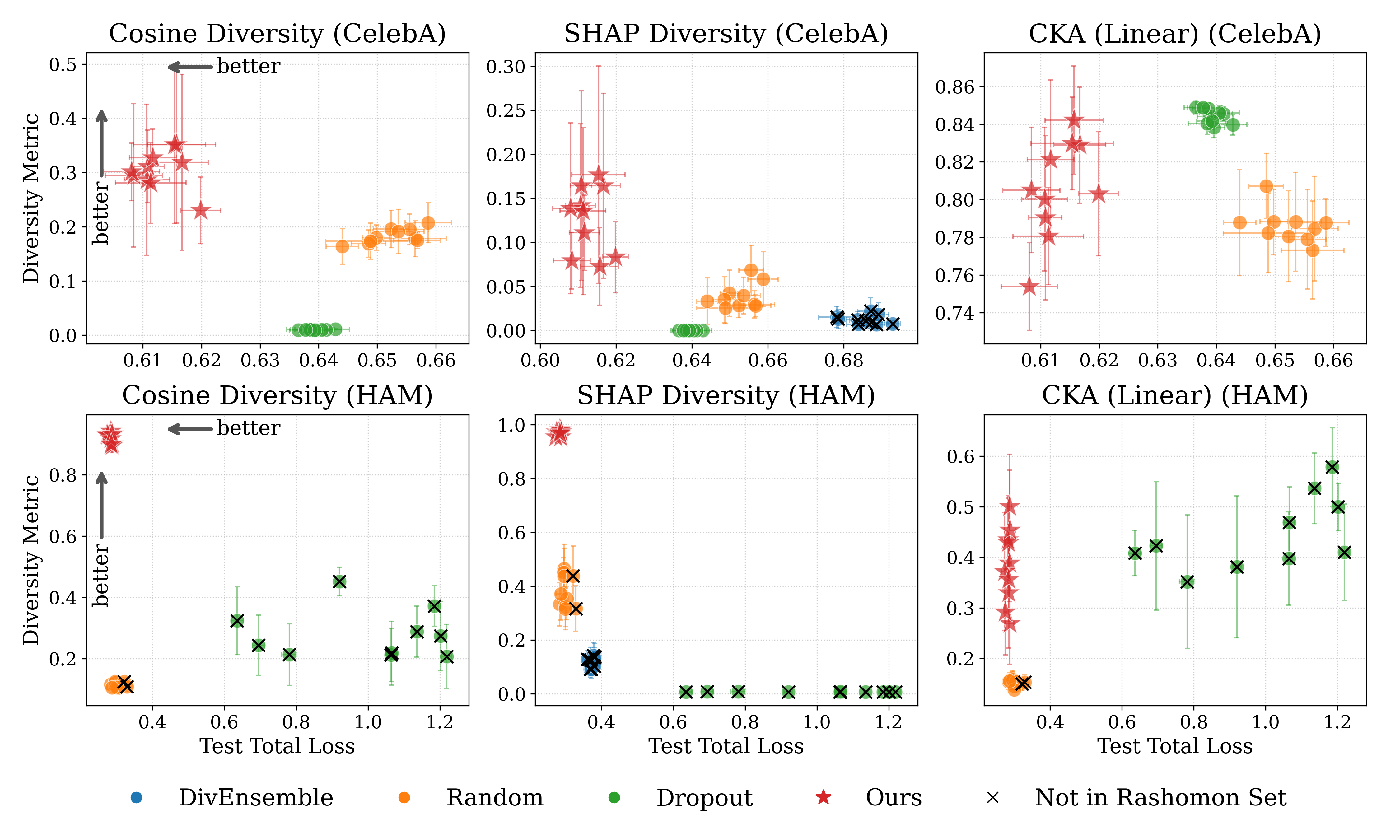}
    \caption{Comparison of model diversity versus total loss on test set across different methods on the CelebA and HAM10000 datasets. The first column displays cosine diversity, the second column displays SHAP diversity, and the last column shows CKA diversity. Each point represents an individual model. Black crosses indicate models that do not satisfy the Rashomon set membership criterion on the validation data.
     Our method achieves higher diversity while remaining within the Rashomon set. Note that Diverse Ensemble baseline is excluded from the plots for the cosine and CKA diversity metrics.  For this baseline, a single concept encoder is shared across all $m$ models, resulting in identical concept representations where diversity metrics are trivially equal to 1.}
    \label{fig:result_ham_celeba}
\end{figure}


\subsection{Diversity Results on More Datasets}\label{appendix:more-diversity-results}

Figure \ref{fig:result_ham_celeba} illustrates the performance of our proposed method compared to baselines on the HAM10000 and CelebA datasets. Similar to what we observed in Figure \ref{fig:task_loss_vs_div}, models found by our method stay in the top-left region for both CelebA and HAM10000, indicating that our method is able to find models that achieve both low test total loss and higher diversity measured by cosine similarity, SHAP, and CKA compared to baselines. Note that because DivEnsemble baseline has a shared concept encoder for all $m$ models, its cosine similarity and CKA similarity are trivially equal to 1. Therefore, we omit these metrics for DivEnsemble.




\section{Efficiency Analysis}\label{app:efficiency}
\begin{table}[htbp]
\centering
\small
\setlength{\tabcolsep}{5pt}
\caption{
Exact memory and parameter costs of different methods and their ratios relative to the random initialization across four datasets.
Our method achieves strong diversity with substantially lower memory and parameter cost than baselines.
Lower is better for all metrics.
}
\resizebox{\linewidth}{!}{%
\begin{tabular}{clrrrr}
\toprule
& Method 
& Mem (GiB) $\downarrow$ & Mem ratio (\%) $\downarrow$
& Params (M) $\downarrow$ & Param ratio (\%) $\downarrow$ \\
\midrule
\multirow{4}{*}{\rotatebox{90}{\textbf{CUB}}}
& random init & 36.01 & - & 217.3 & - \\
& dropout     &  4.15 & 11.5\% &  21.8 & 10.0\% \\
& Diverse Ensemble         &  3.75 & 10.4\% &  21.9 & 10.1\% \\
& Ours        &  \textbf{2.81} & \textbf{7.8\%} &   \textbf{2.9} & \textbf{1.3\%} \\
\midrule
\multirow{4}{*}{\rotatebox{90}{\textbf{CIFAR-10}}}
& random init & 37.09 & - & 217.2 & - \\
& dropout     &  4.16 & 11.2\% &  21.8 & 10.0\% \\
& Diverse Ensemble         &  3.75 & 10.1\% &  21.7 & 10.0\% \\
& Ours        &  \textbf{2.81} & \textbf{7.6\%} &   \textbf{2.9} & \textbf{1.3\%} \\
\midrule
\multirow{4}{*}{\rotatebox{90}{\textbf{AwA2}}}
& random init & 37.09 & - & 217.0 & - \\
& dropout     &  4.16 & 11.2\% &  21.7 & 10.0\% \\
& Diverse Ensemble         &  4.66 & 12.5\% &  21.7 & 10.0\% \\
& Ours        &  \textbf{2.81} & \textbf{7.6\%} &   \textbf{2.7} & \textbf{1.2\%} \\
\midrule
\multirow{4}{*}{\rotatebox{90}{\textbf{CelebA}}}
& random init & 37.09 & - & 216.7 & - \\
& dropout     &  4.16 & 11.2\% &  21.7 & 10.0\% \\
& Diverse Ensemble         &  3.75 & 10.1\% &  21.7 & 10.0\% \\
& Ours        &  \textbf{2.80} & \textbf{7.6\%} &   \textbf{2.3} & \textbf{1.1\%} \\\midrule
\multirow{4}{*}{\rotatebox{90}{\textbf{HAM}}}
& random init & 27.19 & - & 875.6 & - \\
& dropout     &  2.80 & 10.3\% &  87.56 & 10.0\% \\
& Diverse Ensemble         &  2.54 & 9.4\% &  87.57 & 10.0\% \\
& Ours        &  \textbf{1.60} & \textbf{5.8\%} & \textbf{12.27} & \textbf{1.4\%} \\
\bottomrule
\end{tabular}
}
\label{tab:mem-vit-full}
\end{table}
\subsection{Comparison of Memory Use and Trainable Parameters}
\label{app:memory}
Table~\ref{tab:mem-vit-full} shows the memory usage and number of trainable parameters for each method across all datasets. Our method requires fewer than 2\% of the trainable parameters compared to the random initialization baseline.
Our method consistently achieves the lowest memory among all methods and remains substantially below that of random initialization. Together with the strong accuracy and diversity results reported earlier, these findings show that our method achieves accuracy, diversity, and efficiency in both memory usage and trainable parameters.

\subsection{Memory Ablations}
\label{app:mem_ablation}
In Section~\ref{sec:efficiency}, we discussed the memory reduction from LoRA and model-axis checkpointing on AwA2 dataset. Our results generalize beyond it to  CUB dataset, as we show in Table \ref{tab:ablation}.
Checkpointing alone reduces peak memory by 82\%, LoRA alone reduces it by 35\%, and combining both reduces it by 93\%. 
These results show that LoRA and checkpointing reduce different components of memory (parameters and optimizer states versus activations) and their effects accumulate. This reduction is architecture-level rather than dataset-specific, and we expect the same trend to hold for other datasets.

\begin{table}[htbp]
\centering
\caption{Peak GPU memory under different combinations of LoRA and checkpointing on CUB. In our method, checkpointing and LoRA are designed to reduce different components of the memory footprint, and combining them gives the largest memory reduction.}
\label{tab:ablation}
\begin{tabular}{lrrrr}
\toprule
Variant & no-LoRA/no-ckpt & no-LoRA/ckpt & LoRA/no-ckpt & \textbf{LoRA/ckpt} \\
\midrule
Mem (GiB) & 37.75 & 6.84 & 24.75 & \textbf{2.81} \\
\% reduced & 0\% & 82\% & 35\% & \textbf{93\%} \\
\bottomrule
\end{tabular}
\end{table}

\subsection{Wall-Clock running time}
\label{app:time}

We further evaluate whether the memory savings in Section~\ref{sec:efficiency} come at the cost of additional per-epoch runtime, comparing wall-clock time and convergence epochs against random initialization.
Table~\ref{tab:runtime} shows that the per-epoch time of our method is comparable to random initialization (e.g., on AwA2, 833s vs. 843s), with moderate overhead on CUB and CIFAR-10 (ratios of 1.35 and 1.36).
Our method requires more epochs to converge, which is consistent with the observation that LoRA-based adaptation trains slower than full fine-tuning \citep{biderman2024lora}. However,  memory is the fundamental bottleneck for exploring the Rashomon slice of CBMs: training 10 randomly initialized models requires 37 GiB and fits at most 12 models on a single L40 GPU, whereas our method can fit at least 50 models. 

\begin{table}[t]
\centering
\caption{Per-epoch running time and the number of epochs to converge of random initialization vs. our method. The ratio is computed as the per-epoch time of our method divided by that of random initialization. Our method takes comparable per-epoch running time, but requires more epochs to converge.}
\label{tab:runtime}
\begin{tabular}{lrrr}
\toprule
\textbf{Metric} & \textbf{CUB} & \textbf{AwA2} & \textbf{CIFAR-10} \\
\midrule
Per-epoch Time: Random & 106s & 843s & 702s \\
Per-epoch Time: Ours        & 143s & 833s & 954s \\ \hline
Ratio (Ours/Random)               & 1.35 & 0.99 & 1.36    \\ \hline
\# Epochs: Random     & 46 & 34 & 61 \\
\# Epochs: Ours       & 251 & 176 & 290 \\
\bottomrule
\end{tabular}
\end{table}


\section{Additional Qualitative Analysis} \label{qualitative_appendix}
\begin{figure}[ht]
    \centering
    \includegraphics[width=\linewidth]{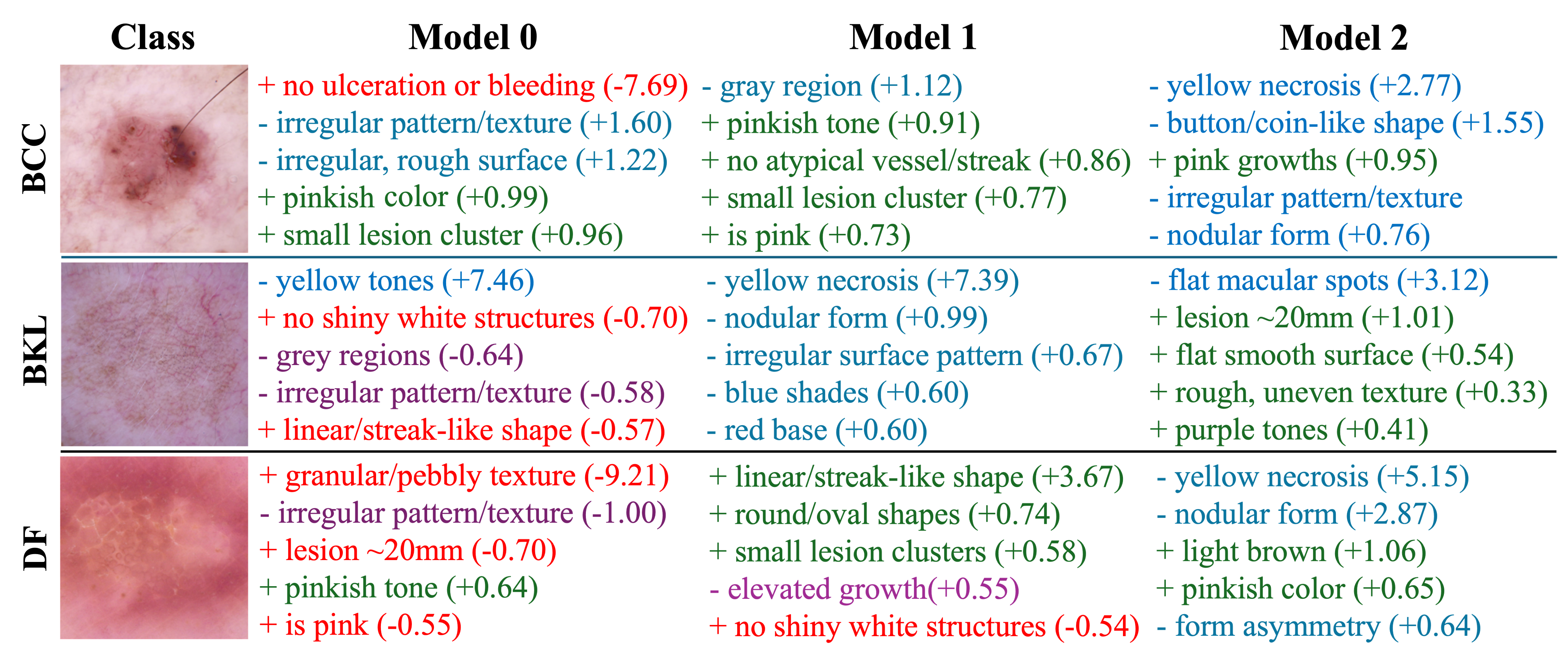}
    \caption{Top-5 concepts based on contribution scores $W_{p,k}\hat{c}_p$ of each concept $p$ to the prediction of class $k$ (value in parentheses) for Basal Cell Carcinoma (BCC, top), Benign Keratosis-like Lesions (BKL, middle), and Dermatofibroma (DF, bottom) classes from the HAM10000 dataset across three Rashomon slice members. The sign stands for whether the concept is predicted to be present ($+$, $\hat{c}_p > 0$) or absent in the corresponding class ($-$, $\hat{c}_p \leq 0$). All the predictions for concepts shown are correct and correspond to ground truth concept labelings for the classes. Color indicates the decision rationale in which the concept contributes to the prediction: \textcolor{teal}{green denotes positive evidence ($W_{p,k}>0, \hat{c}_p>0$);} \textcolor{red}{red represents negative evidence ($W_{p,k}\leq 0, \hat{c}_p>0$);} \textcolor{cyan}{blue means excluding evidence ($W_{p,k}>0, \hat{c}_p\leq0$);} \textcolor{violet}{purple stands for spurious evidence ($W_{p,k}\leq 0, \hat{c}_p\leq 0$).}
    The diverse patterns highlight that different members rely on different rationales.}
    \label{fig:qual_anal_HAM10000}
\end{figure}

\text
In this section, we provide additional qualitative analysis on the HAM10000 dataset. Figure \ref{fig:qual_anal_HAM10000} visualizes three arbitrary models from the Rashomon slice for the prediction of \textbf{Basal Cell Carcinoma (BCC)}, \textbf{Benign Keratosis-like Lesion (BKL)}, and \textbf{Dermatofibroma (DF)} classes. The results are similar to those from the AwA2 dataset (see Figure \ref{fig:qual_anal_AwA2}): different models in the Rashomon slice use distinct combinations of concepts while arriving at the same accurate diagnosis.

Taking dermatofibroma (DF, bottom) as an example, Model 0 is primarily driven by negative evidence.  It identifies texture trait ``granular/pebbly texture'' and size ``approximately 20mm'', but assigns them negative weights, treating these present concepts as counter-evidence of  DF class. Only ``pinkish tone'' serves as direct positive support. 
In contrast, Model 1 relies primarily on positive shape concepts such as ``linear/streak-like shape'', ``round/oval shapes'', and ``small lesion clusters'', with a single exclusion cue (absence of ``elevated growth'') and negative evidence (``no shiny white structures''). Model 2 relies on exclusion reasoning. The absence of pathological features such as ``yellow necrosis'', ``nodular form'', and ``form asymmetry'' serves as the core reasoning, complemented by positive influence of ``light brown'' and ``pinkish color''. 
Overall, the three members span negative-evidence-dominated, positive-shape-driven, and exclusion-based rationales for the DF diagnosis, illustrating that equivalent predictions within a Rashomon slice can emerge from fundamentally different concept-level mechanisms.

\section{More Results on Abstention Use Case}
\label{appendix:usecase}
In cases when CBM is not confident, prior work introduced conceptual safeguards \citep{joren2024classification} that abstain from predicting on high-uncertainty samples and instead seek human intervention to confirm key concepts. 
Traditional safeguards rely on Monte Carlo sampling from a single model to quantify uncertainty. However, this often reflects local variation around a single rationale. A Rashomon set provides a more natural framework for this task, as it contains multiple, near-optimal, but conceptually distinct rationales. 

Therefore, we replace Monte Carlo samples with the Rashomon slice members, treating their disagreement on task predictions as a direct uncertainty signal. When the disagreement rate exceeds the threshold $\tau$, the system abstains and queries an expert for up to $B$ concept values. We then reweight members by their consistency with the expert's 
confirmed concepts:
\[w_m \propto \exp\left(-\mathcal{L}_{C}^{(S)}(g_m(\mathbf{x}), 
\mathbf{c})\right),\]
where $S$ is the set of queried concepts and $\mathbf{c}$ is the 
expert-confirmed concept vector. We accept a prediction when the 
weighted consensus exceeds the threshold:
\[\max_y \sum_m w_m\,\mathbbm{1}_{[f_m(g_m(\mathbf{x}))=y]} \geq \tau,\]
and otherwise query further concepts or abstain once budget $B$ is 
exhausted.

\begin{figure}[t]
    \centering
    \includegraphics[width=\linewidth]{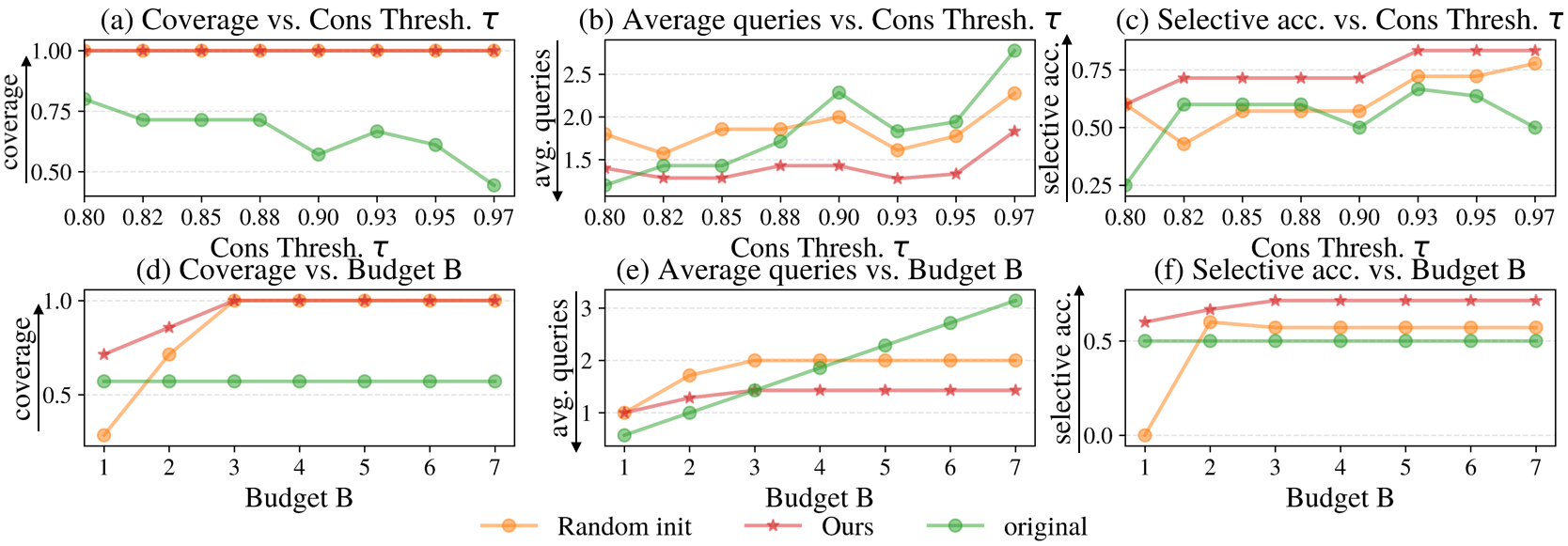}
    \caption{Performance of abstention on beaver class from the AwA2 dataset across varying the confidence thresholds $\tau$ (top row) and query budgets $B$ (bottom row) on the coverage (fraction of non-abstained predictions), average number of concept queries, and selective accuracy (accuracy on non-abstained samples). Arrows indicate the preferred direction. The Rashomon-member safeguard achieves higher coverage with fewer queries while improving selective accuracy across settings compared to baselines.}
    \label{fig:safeguard_AwA2_class3}
\end{figure}

In addition to an example shown in Figure \ref{fig:safeguard_HAM10000_class3}, we compare our Rashomon-based safeguard against both a single-CBM safeguard and the random initialization baseline on Beaver class from the Awa2 dataset in Figure \ref{fig:safeguard_AwA2_class3} across various budgets $B$ and thresholds $\tau$. 
Our method returns a prediction on a larger fraction of test examples without needing to abstain, even at high confidence thresholds (left column). Our models require fewer human concept queries to reach a confident consensus (middle column), and for the samples where a prediction is provided, the accuracy is higher than the baselines (right column). Note that  Diverse Ensemble can not be applied in this use case because all child models in Diverse Ensemble share the same encoder and output the same concept predictions. Our results indicate that diverse Rashomon members can provide a more precise indicator of uncertainty than sampling from a single model.

\section{Diversity Within the Rashomon Slice Members is Meaningful}
\label{app:source_of_diversity}



As discussed in the main text, a practically useful Rashomon slice requires models that exhibit diverse reasoning pathways. To further verify the diversity of our models, we first use concept interventions to reveal the large variance in the models' final decision logic across different Rashomon slice members and then use layer-wise eigenvector visualizations to trace the origin of this diversity to the LoRA modules.

We first demonstrate that members of our Rashomon slice have distinct dependencies on visual features. We perform interventions on the tiger class in AwA2 using eight concepts, specifically: ``orange'', ``stripes'', ``furry'', ``big'', ``paws'', ``tail'', ``meatteeth'', ``claws''. For each model $m$, we intervene on subsets of five concepts at a time by changing the signs of the predicted concept logits and record the resulting accuracy drop over all testing data points belonging to the tiger class. We chose to intervene on five concepts at a time to ensure perturbations are strong enough to produce measurable accuracy differences across models, while leaving enough concepts intact so that models relying on different subsets can respond distinctly.

For a fixed set of concepts we intervene on, the change in accuracy varies wildly across different models $m$: some models display little or no accuracy change (0\% change in accuracy), while others suffer near complete degradation (losing up to 99.5\% accuracy). From these results we find that different models in the Rashomon slice can respond differently to the same concept intervention indicating different reasoning pathways exist. 

To localize the source of diversity in concept representations, we follow \citet{muhlematter2024lora} and quantify weight-space variation via Singular Value Decomposition (SVD).
For each transformer block, we compute the cosine similarity between the top-16 right singular vectors of the LoRA-augmented projection matrices across Rashomon slice members.
We perform this analysis for the QKV and output projection matrices. 

\begin{figure}[htbp]
    \centering \includegraphics[width=\linewidth]{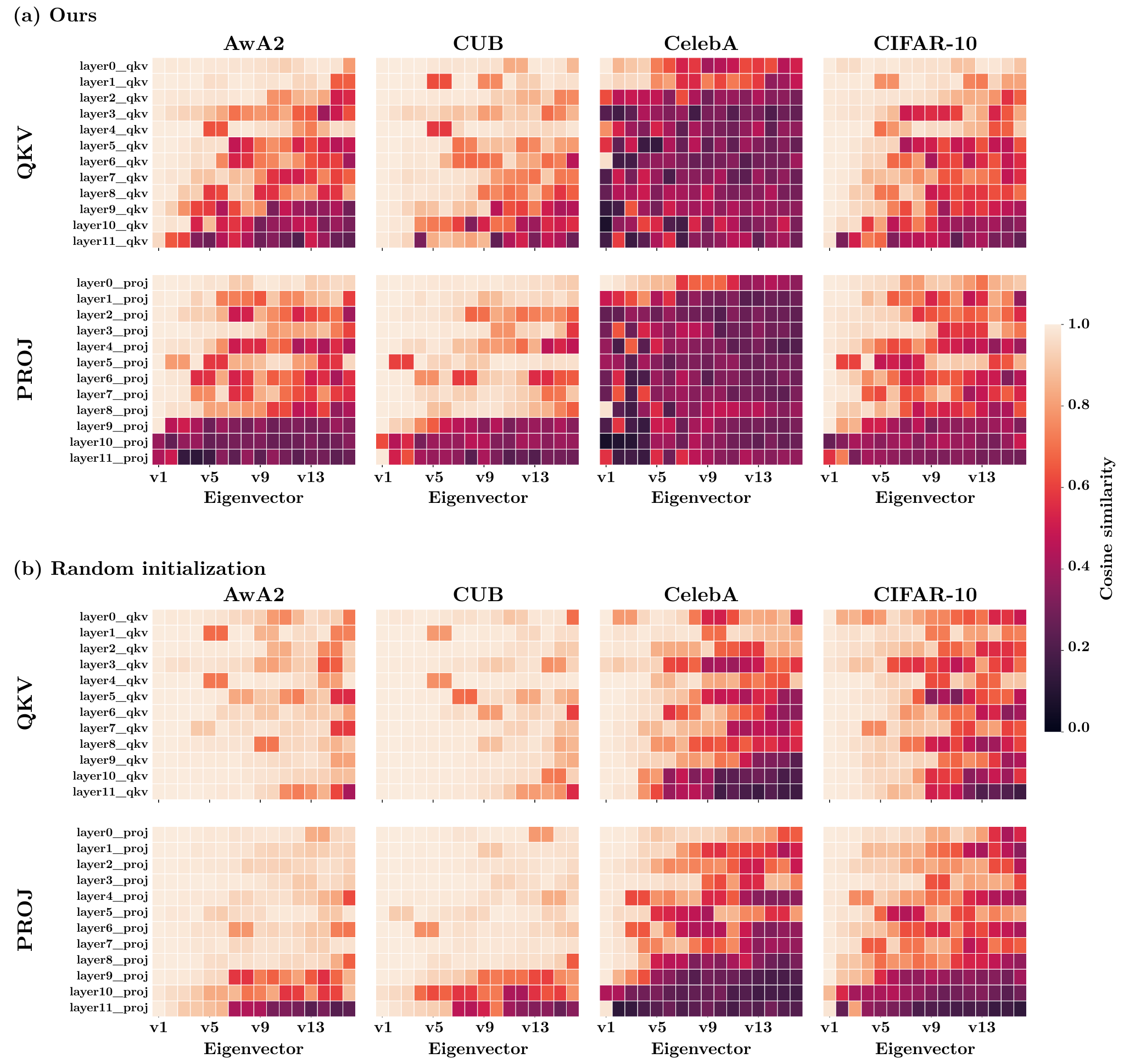}
     \caption{Layerwise eigenvector similarity calculated on the QKV and projection matrices for (a) our Rashomon slice and (b) the random initialization baseline. For the Rashomon slice, eigenvector similarity generally decreases with depth, with lower similarity more prominent in intermediate and deeper layers. Compared with random initialization, the Rashomon slice exhibits lower similarity across a broader range of intermediate and deeper blocks on AwA2, CUB, and CelebA dataset, indicating stronger parameter-space separation among members.}
     \label{fig:Eigen_diversity}
\end{figure}



As shown in Figure~\ref{fig:Eigen_diversity}(a), the leading directions in early layers remain similar across models, while similarity gradually declines with depth and diversity concentrates in the LoRA modules of the deeper blocks.
This means that models can learn low-level concepts in the same way, but diverge when learning higher-level concepts, which explains why our framework can achieve a good balance between accuracy and diversity.

Across datasets and QKV and projection matrices, for the Rashomon slice we observe a clear pattern: similarity decreases noticeably with depth, and within each layer the later eigenvectors show lower similarity than the leading ones. Compared with the random initialization baseline (see Figure~\ref{fig:Eigen_diversity}(b)), our Rashomon slice exhibits stronger separation in several intermediate and deeper blocks in AwA2, CUB, and CelebA. This indicates that Rashomon models share similar low-level representations but diverge increasingly in deeper layers. This structured reduction in similarity reflects the effect of our method in encouraging meaningful representational diversity by using the adaptation modules.

\section{Diversity Penalty and Concept Leakage}
\label{app:leakage}

Since our diversity penalty encourages members to differ in their continuous concept predictions, a potential concern is that members could achieve this diversity by encoding information beyond the corresponding ground-truth concepts, thereby increasing concept leakage. 
We test whether this occurs on the AwA2 dataset.

We compare our Rashomon slice against a baseline trained without the diversity term (Random Init with LoRA). To compare leakage at similar predictive performance, we compare members satisfying the same Rashomon set criterion in Definition~\ref{def:rset}, with $\varepsilon=0.03$.
We use the concept-task leakage (CTL) and interconcept leakage (ICL) scores of \citet{parisini2025leakage}. 
CTL measures whether a predicted concept contains more information about the task label than its corresponding ground-truth concept, while ICL measures whether two predicted concepts are more strongly dependent than the corresponding pair of ground-truth concepts. 
Both are non-negative, with lower values indicating less leakage.

Mathematically, let $c_i$ and $\hat c_i$ denote the ground-truth and predicted values of concept $i$, respectively, and let $y$ denote the task label. Using $I(\cdot;\cdot)$ and $H(\cdot)$ to denote mutual information and entropy, respectively,
\begin{equation}
\mathrm{CTL}_i = \max\!\Big(0,\; \tfrac{I(\hat{c}_i; y)}{H(y)} - \tfrac{I(c_i; y)}{H(y)}\Big),
\end{equation}
\begin{equation}
\mathrm{ICL}_{ij} = \max\!\Big(0,\;
\tfrac{I(\hat{c}_i; \hat{c}_j)}{\sqrt{H(\hat{c}_i)H(\hat{c}_j)}} -
\tfrac{I(c_i; c_j)}{\sqrt{H(c_i)H(c_j)}}\Big).
\end{equation}

We estimate mutual information and entropy using $k$-nearest-neighbor estimators on the held-out test set for each member.

\begin{figure}[htbp]
    \centering
    \includegraphics[width=0.85\linewidth]{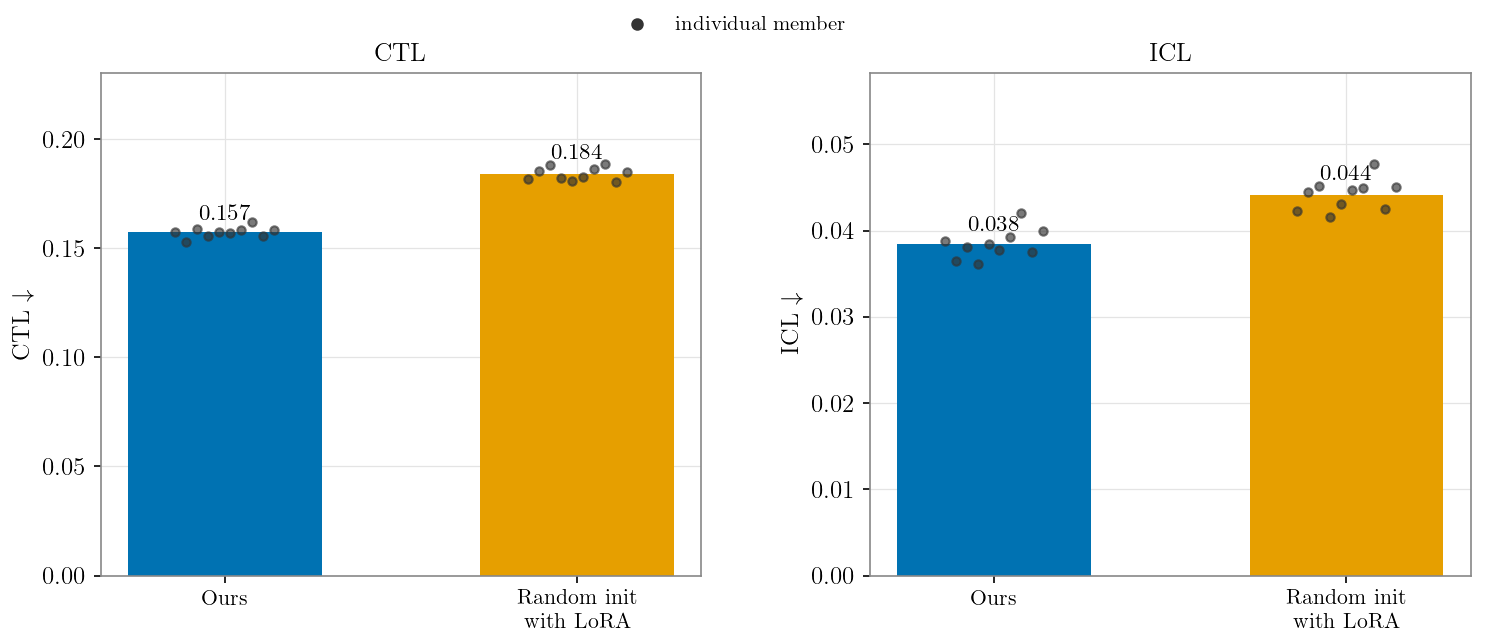}
    \caption{Concept leakage on AwA2. Each point represents one of the $M=10$ members, and each bar shows the mean across members. Lower values indicate less leakage. Our slice has lower mean CTL ($0.157$ vs.\ $0.184$) and ICL ($0.038$ vs.\ $0.044$) than the control baseline trained without the diversity penalty. }
    \label{fig:leakage_per_model}
\end{figure}

As shown in Figure~\ref{fig:leakage_per_model}, our slice has slightly lower average CTL and ICL than the baseline. The per-member scores show a consistent pattern, indicating that the average results are not driven by only a subset of members. This suggests that the increased concept diversity is not accompanied by higher leakage under either metric. 



\section{Sensitivity Analysis of the Rashomon Slice Size}

\begin{figure}[ht]
    \centering
    \includegraphics[width=\linewidth]{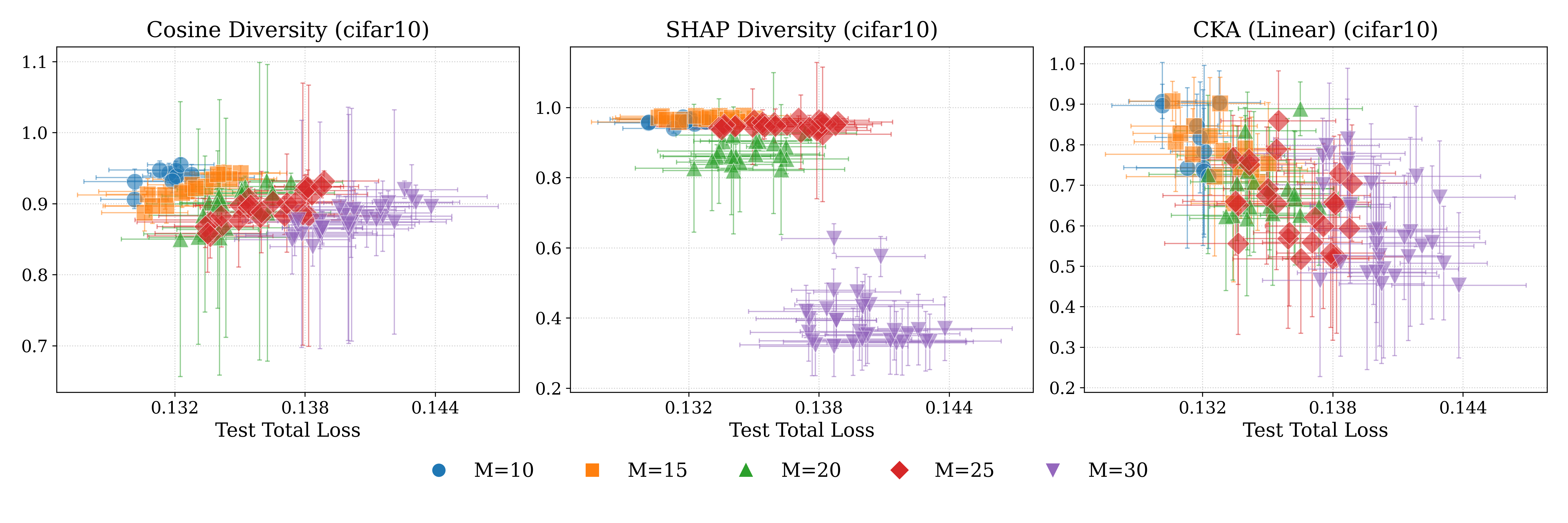}
    \caption{Impact of Rashomon slice size $M \in \{10, 15, 20, 25, 30\}$ on CIFAR-10.  
    As $M$ increases, the test loss for the Rashomon slice members shifts slightly higher. While cosine diversity remains stable, both SHAP and CKA diversity have a visible decrease as $M$ approaches 30. 
    All the models shown are verified members in the Rashomon set.
}
    \label{fig:cifar10_num_models}
\end{figure}

In this section, we study the influence of the Rashomon slice size on the model diversity and accuracy. We vary $M \in \{10, 15, 20, 25, 30\}$ for the CIFAR-10 dataset and report diversity metrics and test total loss in Figure \ref{fig:cifar10_num_models}. 

As we can see, the test total loss tends to increase for models obtained at larger values of $M$. The test total loss for models obtained by $M=30$ (purple triangles) is shifted toward the right compared to smaller slices. This suggests that as we increase the required number of models, the search process must ``reach further'' into the loss landscape, incorporating candidates that are further from the global optimum. 

The impact of increasing $M$ from 10 to 30 varies significantly across the different diversity metrics. While cosine diversity (left) remains relatively stable, indicating that the fundamental parameter orientations are robust to scaling, SHAP diversity and CKA (middle and right) show more sensitivity. Specifically, for $M=30$, there is a visible decreasing trend in diversity scores. This indicates that as the size of the Rashomon slice grows, it becomes increasingly difficult to maintain high pairwise diversity, as additional models begin to converge toward a more redundant latent subspace.

Collectively, these results reveal a trade-off between the size of the Rashomon slice and individual model performance. While a smaller size (e.g., $M=10, 15$) yields a tighter cluster with lower test loss and higher diversity, increasing $M$ beyond 25 could lead to ``representation saturation''. The divergence between weight-based (Cosine) and representation-based (CKA/SHAP) metrics shows that while the models remain distinct in parameter space, their functional behavior becomes increasingly coupled at higher $M$. In practice, an optimal configuration must balance the quantity of models discovered with the meaningful diversity and accuracy of each member. 

\section{Sensitivity Analysis of \texorpdfstring{$\alpha$}{alpha}}

To characterize the relationship between diversity and concept accuracy 
across values of the regularization strength $\alpha$ in 
Eq.~\ref{eq:total_loss}, we conduct a grid search over $\alpha$ on the 
AwA2 and CUB datasets using $M=5$ models, with all other hyperparameters and training process the same as in Section~4. We evaluate the effect of varying $\alpha$ on concept accuracy and three diversity metrics: cosine similarity, SHAP diversity, and CKA. Task accuracy remains stable across models and values of $\alpha$ due to the minimax objective in Eq.~\ref{eq:total_loss}, so the analysis focuses on the interplay between concept accuracy and diversity.

\begin{figure}[t]
    \centering
    \includegraphics[width=0.99\linewidth]{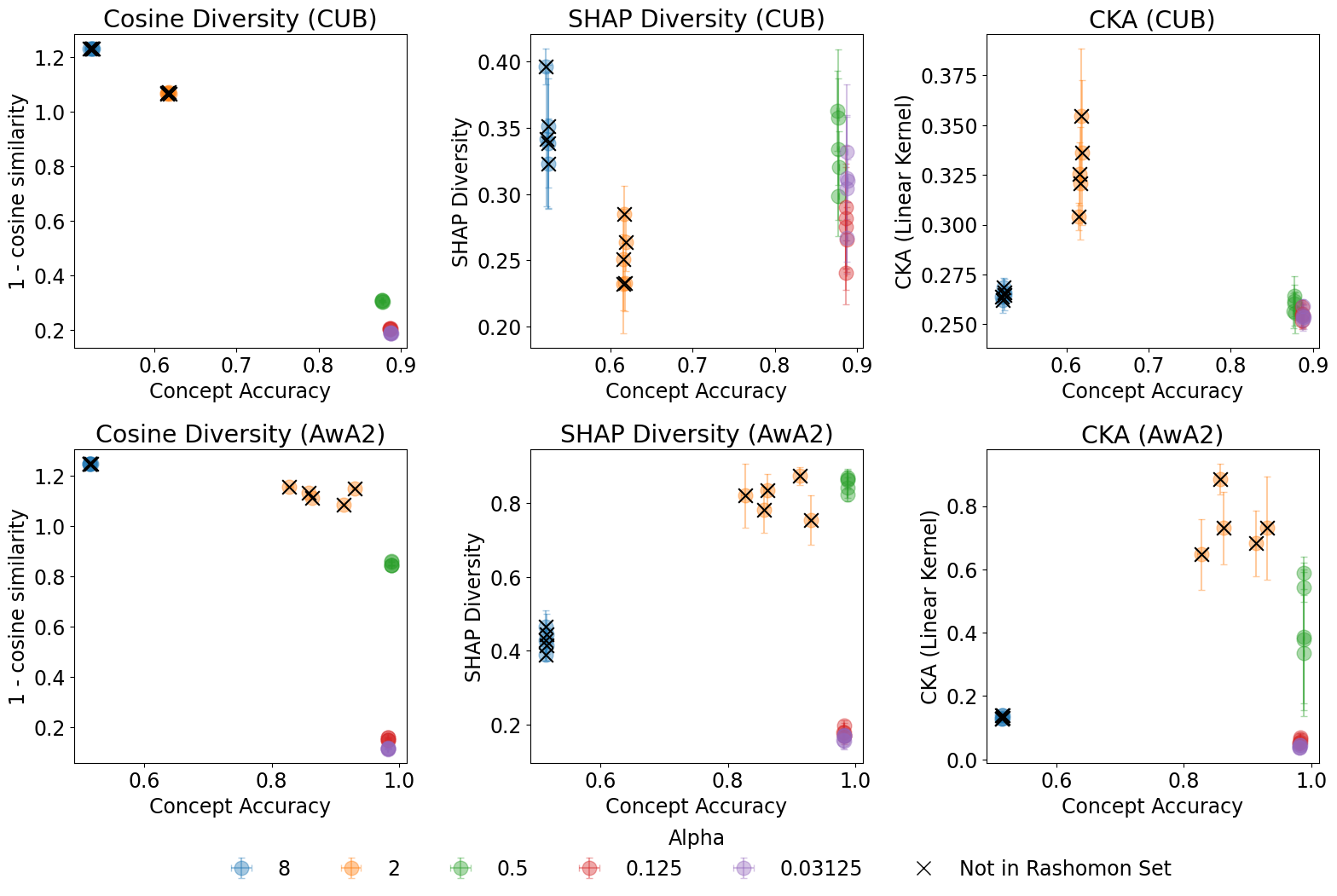}
    \caption{Diversity measures versus concept accuracy for different values 
    of $\alpha$ on CUB (top) and AwA2 (bottom). Crosses indicate models outside the 
    Rashomon set. The relationship between diversity and concept accuracy 
    is metric- and dataset-dependent, with CUB showing a sharper frontier 
    under cosine distance than AwA2.}
    \label{fig:diversity_concept_tradeoff}
\end{figure}

The results are presented in Figure~\ref{fig:diversity_concept_tradeoff}. Increasing $\alpha$ generally improves diversity at some cost to concept accuracy, though the sharpness of this trade-off varies by dataset and metric. On CUB, the frontier is sharper under cosine distance, with high 
$\alpha$ values risking Rashomon set membership, while on AwA2 the trade-off is softer and higher diversity can often be achieved with little cost to concept accuracy. Overall, in our experiments, values of $\alpha \approx 0.5$ generally provided 
a good balance between diversity and concept accuracy while keeping 
models within the Rashomon set.  This motivates our dynamic $\alpha$ 
schedule, which is initialized around this range and adapts during training without requiring expensive per-dataset grid searches.

\section{Optimization Stability}
\begin{figure}[t]
\centering
\includegraphics[width=\linewidth]{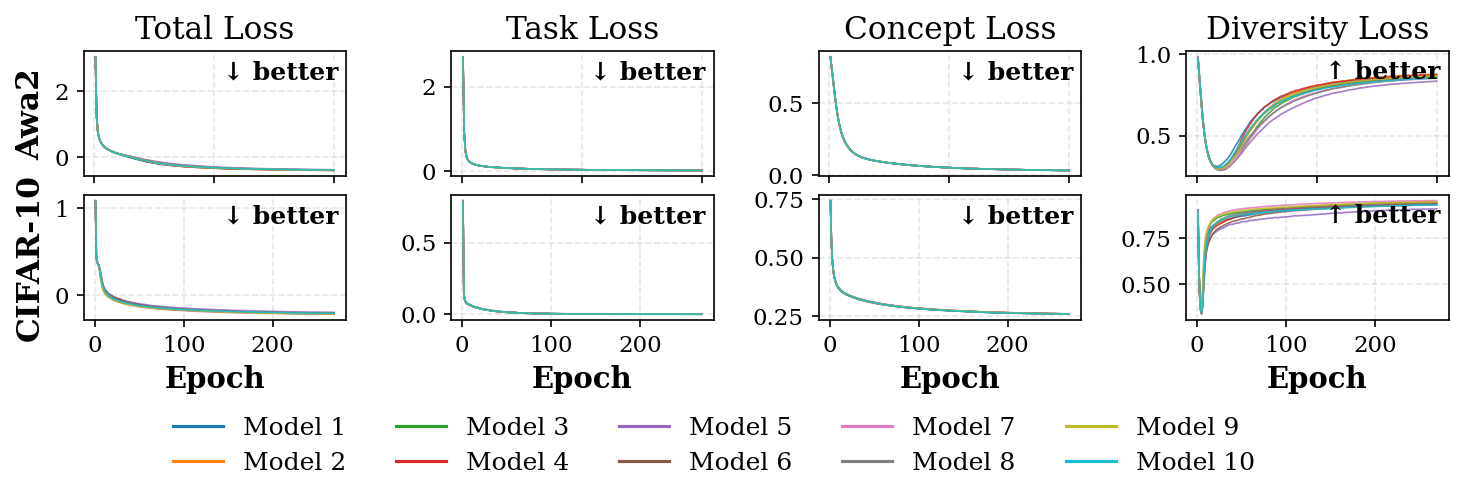}
\caption{Training dynamics of all Rashomon slice members during optimization on CIFAR-10 and AwA2. Total, task, concept, and diversity losses all converge smoothly without oscillation or collapse, indicating stable training in practice.}
\label{fig:optimization_stability}
\end{figure}

In this section, we report the training curves of all Rashomon slice members during optimization. Figure~\ref{fig:optimization_stability} shows that the training losses of all members converge smoothly with no oscillation or collapse, confirming that our optimization remains stable in practice.

\end{document}